\documentclass{article}
\usepackage[top=2.5cm,bottom=2.5cm,left=2.5cm,right=2.5cm]{geometry}
\usepackage{common}

\usepackage[
bookmarks, colorlinks, breaklinks, 
]{hyperref}  
\hypersetup{linkcolor=darkmidnightblue,citecolor=darkmidnightblue,filecolor=darkmidnightblue,urlcolor=darkmidnightblue} 
\usepackage{graphicx}
\makeatletter
\def\blfootnote{\xdef\@thefnmark{}\@footnotetext}
\makeatother
\usepackage{amsmath}
\usepackage{amsmath,textcomp,amssymb,geometry,graphicx,enumerate}
\usepackage{algorithm}
\usepackage{stmaryrd}
\usepackage{parskip}
\usepackage{tikz}
\usepackage{natbib}

\usetikzlibrary{fit,positioning}
\usetikzlibrary{bayesnet}
\usetikzlibrary{arrows}

\begin{document}
\definecolor{darkmidnightblue}{rgb}{0.0, 0.25, 0.45}
\title{ \Large A Tutorial on Deep Latent Variable Models of Natural Language} 

\author{Yoon Kim$^\ast$ \\ 
{\normalsize \texttt{ yoonkim@seas.harvard.edu}} \\ 
{\normalsize Department of Computer Science} \\ {\normalsize  Harvard University} 
\and Sam Wiseman$^\ast$ \\
{\normalsize \texttt{ swiseman@ttic.edu}} \\ 
{\normalsize Toyota Technological}  \\ {\normalsize  Institute at Chicago} 
\and Alexander M. Rush \\ 
{\normalsize \texttt{ srush@seas.harvard.edu}} \\ 
{\normalsize Department of Computer Science} \\ {\normalsize  Harvard University} }

\date{}
\maketitle
\blfootnote{$^\ast$Equal contribution.}
\textit{This manuscript was written to complement a tutorial at EMNLP 2018 and is 
intended to serve both as an introduction to deep latent variable models as well as a review of recent advances. Its current state is somewhere between a tutorial and notes. The accompanying slides and code can be found at \url{http://nlp.seas.harvard.edu/latent-nlp-tutorial.html} }
\section{Introduction}
\label{sec:intro}
\emph{Latent variable models}, and probabilistic graphical models more generally, provide a declarative language for specifying prior knowledge and structural relationships in complex datasets. They have a long and rich history in natural language processing, having contributed to fundamental advances such as statistical alignment for translation \citep{brown1993mt}, topic modeling~\citep{blei2003latent}, unsupervised part-of-speech tagging \citep{brown1992class}, and grammar induction \citep{klein2004corpus}, among others. 
\emph{Deep learning}, broadly construed, is a toolbox for learning rich representations (i.e., features) of data through numerical optimization. Deep learning is the current dominant paradigm in natural language processing, and some of the major successes include language modeling \citep{Bengio2003,Mikolov2010,Zaremba2014}, machine translation \citep{Sutskever2014,Cho2014,Bahdanau2015,vaswani2017attention},  and natural language understanding tasks such as question answering and natural language inference.

There has been much recent, exciting work on combining the complementary strengths of latent variable models and deep learning. Latent variable modeling makes it easy to explicitly specify model constraints through conditional independence properties, while deep learning makes it possible to parameterize these conditional likelihoods with powerful function approximators. While these ``deep latent variable" models provide a rich, flexible framework for modeling many real-world phenomena, difficulties exist:
deep parameterizations of conditional likelihoods usually make posterior inference intractable, and latent variable objectives often complicate backpropagation by introducing points of non-differentiability. This tutorial explores these issues in depth through
the lens of \emph{variational inference} \citep{Jordan1999,Wainwright2008}, a key technique for performing approximate
posterior inference.

The term ``deep latent variable" models can also refer to the use of neural networks to perform latent variable \textit{inference } (``deep inference"). In the context of variational inference, this means that we train an inference network to output the parameters of an approximate posterior distribution given the set of variables to be conditioned upon \citep{Kingma2014,Rezende2014,Mnih2014}.  We will devote a significant portion of the tutorial to this setting.

\paragraph{Tutorial Outline}
The tutorial will be organized as follows: section~\ref{sec:intro} introduces notation and briefly describes the basics of neural networks; section~\ref{sec:models} presents
several archetypical latent variable models of text and their ``deep" variants; section~\ref{sec:motivation} surveys applications of latent variable models introduced in the previous section; section~\ref{sec:learningin} deals with learning and
performing posterior inference in latent variable models, both in the case where exact inference over the latent variables is tractable and when it is not; section~\ref{sec:deepinf} focuses on
amortized variational inference and variational autoencoders, a central
framework for learning deep generative models; section~\ref{sec:othermethods}
briefly touches on other methods for learning latent variable models; and finally, section~\ref{sec:discussion} concludes.

While our target audience is the community of 
natural language processing (NLP) researchers/practioners,
we hope that this tutorial will also be of use to researchers in other areas. We have therefore organized the tutorial to be modular: sections~\ref{sec:models} and~\ref{sec:motivation}, which are more specific to NLP, are largely independent from sections~\ref{sec:learningin} and~\ref{sec:deepinf}, which mostly deal with the general problem of learning deep generative models. 

\paragraph{Scope}
This tutorial will focus on learning latent variable models whose joint distribution
can be expressed as a \emph{directed graphical model} (DGM),\footnote{Directed graphical models are also sometimes referred to as \emph{generative models}, since their directed nature makes it easy to generate data through ancestral sampling.} and we will
mostly do this through variational inference. Specifically, we will not cover (or only briefly touch on) undirected graphical models such as restricted Boltzmann Machines (and more broadly, Markov random fields), posterior inference based on Markov chain Monte Carlo sampling, spectral learning of latent variable models, and non-likelihood-based approaches such as generative adversarial networks \citep{goodfellow2014generative}. While each of these topics is a rich area of active research on its own, we have chosen to
limit scope of this tutorial to directed graphical models and variational inference
in order to provide a more thorough background on fundamental ideas and key techniques, as well as a detailed survey of recent advances.

\section{Preliminaries}
\label{sec:intro}
\paragraph{Notation}
Throughout, we will assume the following notation:
\begin{itemize}
\item $x$: an observation, usually a sequence of tokens,  $x = x_1, \ldots, x_T$, sometimes denoted with $x_{1:T}$
\item $x^{(1:N)}$: a dataset of $N$ i.i.d observations, $x^{(1:N)} = x^{(1)}, \dots, x^{(N)}$
\item $z$: an unobserved latent variable, which may be a sequence or other structured object
\item $\boldz$: a latent vector (we will use $z$ to denote a general latent variable, and  $\boldz$ in the specific case where the latent variable is a continuous vector, e.g. $\boldz \sim \mcN(\mathbf{0}, \mathbf{I})$)
\item  $\theta$:  generative model parameters
\item $\lambda$: variational parameters

\item $p(x,z \param \theta)$, or $p_{x,z}(x,z \param \theta)$: generative model parameterized by $\theta$
\item $p(x \given z \param \theta)$, or $p_{x |z}(x \given z \param \theta)$: likelihood model parameterized by $\theta$
\item $q(z \param \lambda)$: variational distribution parameterized by $\lambda$
\item $\Delta^{V-1}$: the standard $V$-simplex, i.e. the set $\{\bpi \in \reals^V \mid \pi_v \geq 0 \text{ for all $v$, and } \sum_{v=1}^V \pi_v = 1 \}$
\end{itemize}
We will use $p(x \param \theta)$ to denote the probability mass (or density) function
of a random variable evaluated at $x$, where the density is parameterized by $\theta$. While the random variable over which the distribution is induced will be often clear from context, when it is not clear we will either use subscripts to identify the random variable, e.g. $p_z(z \param\theta)$ and $p_x(x \param \theta$), or a different letter, e.g. $q(z \param \lambda)$.
We will sometimes overload notation and use $p(x = x_0 \param \theta)$ with the same $p$ to refer to the probability of the event that the random variable $x$ takes on the value $x_0$ where the probability distribution is again parameterized by $\theta$, i.e. $p(x = x_0 \param \theta) = p_x(x_0 \param \theta)$. For brevity of
notation, and since the distinction will encumber rather than elucidate at the level of abstraction this tutorial is aiming for, we will also
overload variables (e.g. $x, z$) to both refer to a random variable (a measurable function from a sample space $\Omega$ to $\reals^d$) and its realization (an element of $\reals^d$). 
\paragraph{Neural Networks}
We now briefly introduce the neural network machinery to be used in this tutorial. Deep networks are parameterized non-linear functions, which transform an input $\boldz$ into features $\boldh$ 
using parameters $\pi$. We will in particular make use of the multilayer perceptron (MLP), which computes features as follows:
\begin{align*}
\boldh = \MLP(\boldz \param \pi) = \boldV \sigma (\boldW  \boldz + \boldb) + \bolda,
\end{align*}
where $\sigma$ is an element-wise nonlinearity, such as $\tanh$, $\RELU$, or the logistic sigmoid, and the set of neural network parameters is $\pi = \{ \boldV, \boldW,  \bolda, \boldb\}$.

We will also make use of the recurrent neural network (RNN), which maps a sequence of inputs $\boldz_{1:T}$ into a sequence of features $\boldh_{1:T}$, as follows:
\begin{align*}
\boldh_{t} =  \RNN(\boldh_{t-1}, \boldz_t \param \pi) = \sigma (\boldU  \boldz_t + \boldV \boldh_{t-1} + \boldb),
\end{align*}
where $\pi = \{ \boldV, \boldU, \boldb\}$, and where for concreteness we have parameterized the above RNN as an Elman RNN \citep{elman1990}. While modern architectures typically use the slightly more complicated long short-term memory (LSTM)~\citep{hochreiter1997long} or gated recurrent units (GRU)~\citep{Cho2014} RNNs instead, we will remain
agnostic with regard to the specific RNN parameterization, and simply use $\RNN(\boldh_{t-1}, \boldz_t \param \pi)$ to encompass the  LSTM/GRU parameterizations as well.

\paragraph{RNN Language Models}
We will shortly introduce several archetypal latent variable models of a sentence $x = x_1, \ldots, x_T$ consisting of $T$ words. Before we do so, however, we briefly introduce the RNN language model, an incredibly effective model of a sentence that uses \textit{no} latent variables. Indeed, because RNN language models are such good sentence models, we will need a very good reason to make use of latent variables, which as we will see complicate learning and inference.
Sections~\ref{sec:models} and ~\ref{sec:motivation} attempt to motivate the need for introducing latent variables into our sentence models, and the remaining sections discuss how to learn and do inference with them.

Using the notation $x_{<t}$ as shorthand for the sequence $x_1, \ldots, x_{t-1}$, RNN language models define a distribution over a sentence $x$ as follows
    \begin{align} \label{eq:rnnlm}
     p(x_{1:T}) = \prod_{t=1}^T p(x_t \given x_{<t}) = \prod_{t=1}^{T} \softmax(\boldW \boldh_t)_{x_{t}},
    \end{align}
where 
\begin{align} \label{eq:rnn}
\boldh_t = \RNN(\boldh_{t-1}, \boldx_{t-1};\theta),
\end{align}
and $\boldx_{t-1}$ is the word embedding corresponding to the $(t-1)$-th word in the sequence. The $\softmax$ function over a vector $\bolds \in \reals^V$ applies an element-wise exponentiation to $\bolds$ and renormalizes to obtain a distribution, i.e. $\softmax: \reals^V \to \Delta^{V-1}, \,\, \boldy = \softmax(\bolds)$ means that each element of $\boldy$ is given by
\[ \boldy_k = \frac{\exp(\bolds_k)}{\sum_{v=1}^V \exp(\bolds_v)}. \]
Because RNN language models are a central tool in deep NLP, for the remainder of the tutorial we will use the notation 
\begin{align*}
x_{1:T} \sim p(x_{1:T})= \RNNLM(x_{1:T} \param \theta)    
\end{align*}
to mean that $x$ is distributed according to an RNN language model (as above) with parameters $\theta$, where $\theta$ includes the parameters of the RNN itself as well as the matrix $\boldW$ used in defining the per-word distributions at each step $t$.

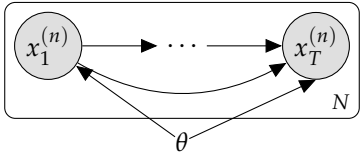
\begin{figure}
\centering
\begin{tikzpicture}
\node (dots) {$\ldots$};%
 \node[obs, left=1cm of dots] (x1) {$x_1^{(n)}$};%
 \node[obs, right=1cm of dots] (xT) {$x_T^{(n)}$};%
 \node[const, below=1cm of dots] (pi) {$\theta$};
 
 \plate {plate1} {(dots)(x1)(xT)} {$N$}; %
 \edge {x1} {dots};
  \edge {dots} {xT};
 \edge {pi} {x1,xT.south};

 \draw[->] 
 (x1) edge[bend right] node [right] {} (xT);
\end{tikzpicture}
 \caption{Graphical model corresponding to an RNN language model.}
 \label{fig:rnnlmgm}
\end{figure}

Finally, we note that the $\RNNLM$ model above makes no independence assumptions, and simply decomposes the probability of the sequence $x$ using the chain rule of probability. We show a graphical model corresponding to an RNN language model in Figure~\ref{fig:rnnlmgm}.

\section{Archetypal Latent Variable Models of Sentences}
\label{sec:models}
We now present several archetypal latent variable models of sentences $x = x_1, \ldots, x_T$, which will serve to ground the discussion. All of the models we introduce will provide us with a joint distribution $p(x, z \param \theta)$ over the observed words $x$ and unobserved latent variables $z$. These models will differ in terms of whether the latent variables $z$ are discrete, continuous, or structured, and we will provide both a shallow and deep variant of each model. We will largely defer the motivating applications for these models to the next section, but we note that on an intuitive level, the latent variables in these three types of models have slightly different interpretations. In particular, discrete latent variables can be interpreted as inducing a clustering over data points, continuous latent variables can be interpreted as providing a dimensionality reduction of data points, and structured latent variables can be interpreted as unannotated structured objects (i.e., objects with interdependent pieces or parts) in the data;\footnote{Because discrete structured latent variables consist of interdependent discrete latent variables, we can also think of them as inducing interdependent clusterings of the parts of each data point.} 
these interpretations will be expanded upon below.

In addition to describing models in each of the three categories listed above, we will briefly sketch out what inference -- that is, calculating the posterior over latent variables $p(z \given x \param \theta)$ -- in these models looks like, as a way of motivating the later sections of the tutorial.

\subsection{A Discrete Latent Variable Model}\label{naivebayes}
\label{sec:catlatmodel}
We begin with a simple discrete latent variable model, namely, a latent-variable Naive Bayes model (i.e., a mixture of categoricals model). This model assumes a sentence $x = x_1, \ldots, x_T$ is generated according to the following process:

\begin{enumerate}
\item Draw latent variable $z \in \{1, \ldots, K\}$ from a Categorical prior $p(z ;\ \bmu)$ with parameter $\bmu \in \Delta^{K-1}$. That is,
$p(z = k ;\ \bmu) = \mu_k$.
\item Given $z$, draw each token $x_t \in \{1, \dots, V\}$ in $x$ independently from a Categorical distribution with parameter $\bpi_{z} \in \Delta^{V-1}$. That is, $p(x_t = v \param \bpi_{z}) = \pi_{z, v}$,
where $\pi_{z, v}$ is the probability of drawing word index $v$ given the latent variable $z$.
Thus, the probability of the sequence $x = x_1, \dots, x_T$ given $z$ is
\[ p(x \given z ; \ \bpi_{z}) = \prod_{t=1}^T \pi_{z, x_t} \; . \]
\end{enumerate}

Letting $\theta = [\bmu, \bpi_{1}, \ldots, \bpi_K]$ be all the parameters of our model, the full joint distribution is
\begin{align} \label{eq:nb}
 p(x, z ; \ \theta) = p(z;\ \bmu) \times p(x \given z; \ \bpi_{z}) = \mu_{z} \times  \prod_{t=1}^T \pi_{z, x_t} \; .
 \end{align}

This model assumes that each token in $x$ is generated independently,
conditioned on $z$. This assumption is clearly naive (hence the name) but greatly reduces the number
of parameters that need to be estimated.\footnote{In our formulation we model text as \emph{bag-of-words} and thus ignore position information. It is also possible to model position-specific probabilities within Naive Bayes with additional parameters $\pi_{z,v,t}$ for $t = 1, \dots T$. This would result in $KVT$ parameters.} The total number of parameters in this generative model is $K + KV$, where we have $K$ parameters for $\mu$ and $V$ parameters in $\pi_z$ for each of the $K$ values of $z$.\footnote{The model is overparameterized since we only need $V-1$ parameters for a Categorical distribution over a set of size $V$. This is rarely an issue in practice.} 

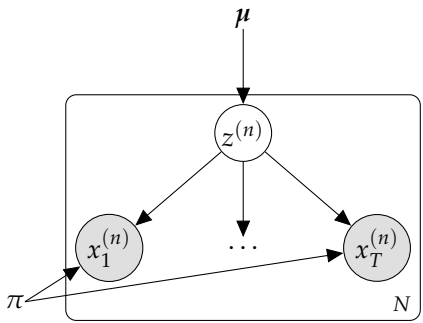
\begin{figure}
\centering
\begin{tikzpicture}
\node (dots) {$\ldots$};%
 \node[obs, left=1cm of dots] (x1) {$x_1^{(n)}$};%
 \node[obs, right=1cm of dots] (xT) {$x_T^{(n)}$};%
 \node[latent, above=of dots] (z) {$z^{(n)}$}; %
 \node[const, above=of z] (mu) {$\bmu$};
 \node[const, below left=0.3cm and 0.8cm of x1] (pi) {$\pi$};
 
 \plate {plate1} {(dots)(x1)(xT)(z)} {$N$}; %
 \edge {z} {dots};
 \edge {z} {x1};
 \edge {z} {xT};
 \edge {mu} {z};
 \edge {pi.east} {x1,xT};
 \end{tikzpicture}
 \caption{Naive Bayes graphical model. For simplicity, all sequences are depicted as having $T$ tokens. All distributions are categorical, and the parameters are $\bmu \in \Delta^{K-1}$ and $\pi = \{\bpi_k \in \Delta^{V-1}\}_{k=1}^K$.}
 \label{fig:nbgm}
\end{figure}

Despite the Naive Bayes assumption, the above model becomes interesting when we have $N$ sentences $\{x^{(n)}\}_{n=1}^N$ that we assume to be generated according to the above process. Indeed, since each sentence $x^{(n)}$ comes with a corresponding latent variable $z^{(n)}$ governing its generation, we can see the $z^{(n)}$ values as inducing a clustering over the sentences $\{x^{(n)}\}_{n=1}^N$; sentences generated by the same value of $z^{(n)}$ belong to the same cluster. We show a graphical model depicting this scenario in Figure~\ref{fig:nbgm}.

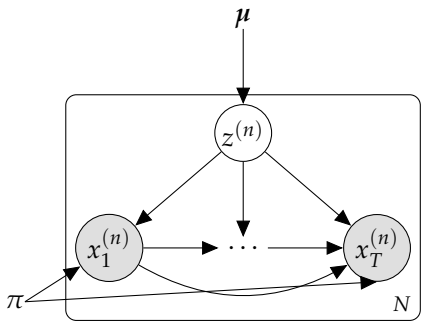
\begin{figure}
\centering
\begin{tikzpicture}
\node (dots) {$\ldots$};%
 \node[obs, left=1cm of dots] (x1) {$x_1^{(n)}$};%
 \node[obs, right=1cm of dots] (xT) {$x_T^{(n)}$};%
 \node[latent, above=of dots] (z) {$z^{(n)}$}; %
 \node[const, above=of z] (mu) {$\bmu$};
 \node[const, below left=0.3cm and 0.8cm of x1] (pi) {$\pi$};
 
 \plate {plate1} {(dots)(x1)(xT)(z)} {$N$}; %
 \edge {z} {dots};
 \edge {z} {x1};
 \edge {z} {xT};
 \edge {mu} {z};
 \edge {x1} {dots};
  \edge {dots} {xT};
 \edge {pi.east} {x1,xT.south};
 
 \draw[->] 
 (x1) edge[bend right] node [right] {} (xT);
 
\end{tikzpicture}
 \caption{Graphical model representation of a categorical latent variable model with tokens generated by an RNN. For simplicity, all sequences are depicted as having $T$ tokens. The $z^{(n)}$s are drawn from a Categorical distribution with parameter $\bmu$, while $x^{(n)}$ is drawn from an $\RNNLM(x \param \pi_{z^{(n)}})$. These $\RNNLM$s have parameters $\pi = \{\pi_k\}_{k=1}^K$.}
 \label{fig:catrnngm}
\end{figure}

\paragraph{Making the Model ``Deep''} One of the reasons we are interested in deep latent variable models is that neural networks make it simple to define flexible distributions without using too many parameters. As an example, 
we can formulate a sentence model similar to the Naive Bayes model, but which avoids the Naive Bayes assumption above (whereby each token is generated independently given $z$) using an RNN. An RNN will allow the probability of $x_{t}$ to depend on the entire history $x_{<t} = x_1, \dots, x_{t-1}$ of tokens preceding $x_{t}$. 
In this deep variant, we might then define the probability of $x$ given latent variable $z$ as 
\begin{align} \label{eq:rnnnb}
p(x \given z) = \RNNLM(x \param \pi_z). 
\end{align}
That is, the probability of $x$ given $z$ is given by an $\RNNLM$ with parameters that are specific to the $z$ that is drawn.
We then obtain the joint distribution $p(x, z \param \theta)$ by substituting~\eqref{eq:rnnnb} into the term for $p(x \given z)$ in~\eqref{eq:nb}. We show the corresponding graphical model in Figure~\ref{fig:catrnngm}. 

Note that this deep model allows for avoiding the Naive Bayes assumption while still using only $O(Kd^2 + Vd)$ parameters, assuming $\boldh_t \in \reals^d$. The Naive Bayes model, on the other hand, requires $O(KV)$ parameters, and $O(KVT)$ parameters if we use position-specific distributions. Thus, as long as $d^2$ is not too large, we can save parameters under the deep model as $K$ and $V$ get large. 

\paragraph{Inference}
For discrete latent variable models, inference -- that is, calculating the posterior $p(z \given x)$ -- can typically be performed by enumeration. In particular, using Bayes's rule we have
\begin{align*}
    p(z \given x \param \theta) = \dfrac{p(x, z \param \theta)}{p(x \param \theta)} = \dfrac{p(z \param \theta) p(x \given z \param \theta)}{\sum_{k=1}^K p(z = k \param \theta) p(x \given z \param \theta)},
\end{align*}
where the latent variable $z$ is assumed to take on one of $K$ values. Calculating the denominator $\sum_{k=1}^K p(z = k \param \theta) p(x \given z \param \theta)$ is clearly the most computationally expensive part of inference, but is generally considered to be tractable as long as $K$ is not too big. Note, however, that the model's parameterization can affect how quickly we can calculate this denominator: under the Naive Bayes model we can accumulate $x$'s word counts once, and evaluate their probability under the $K$ categorical distributions, whereas for the RNN based model we need to run $K$ different RNNs over $x$.

\subsection{A Continuous Latent Variable Model}\label{reallatent}
\label{sec:reallatmodel}
\begin{figure}
\centering
\begin{tikzpicture}
\node (dots) {$\ldots$};%
 \node[obs, left=1cm of dots] (x1) {$x_1^{(n)}$};%
 \node[obs, right=1cm of dots] (xT) {$x_T^{(n)}$};%
 \node[latent, above=of dots] (z) {$\boldz^{(n)}$}; %
 \node[const, above=of z] (mu) {$\bmu$};
 \node[const, below left=0.3cm and 0.8cm of x1] (pi) {$\pi$};
 
 \plate {plate1} {(dots)(x1)(xT)(z)} {$N$}; %
 \edge {z} {dots};
 \edge {z} {x1};
 \edge {z} {xT};
 \edge {mu} {z};
 \edge {pi.east} {x1,xT};
 
 
\end{tikzpicture}
 \caption{Continuous Naive Bayes model. The $\boldz^{(n)}$ have a normal distribution $\mcN(\bmu, \ident)$, and each token $x^{(n)}_t$ has a Categorical distribution with parameter $\softmax(\boldW \boldz^{(n)})$. (For consistency with the previous models, we let $\pi = \{\boldW\}$). Note that the dependence structure is identical to that in Figure~\ref{fig:nbgm}; the only difference is the type of latent variable and the parameterizations.}
 \label{fig:realnbgm}
\end{figure}
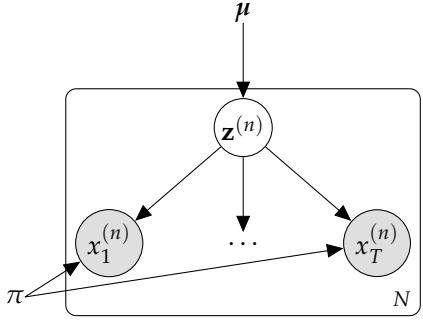

We now consider models in which the latent variables are vectors in $\reals^d$, rather than integer-valued.
We begin with a continuous analog of the Naive Bayes model in the previous subsection. In particular, we will assume a sequence is generated according to the following process:
\begin{enumerate}
    \item Draw latent variable $\boldz$ from $\mcN(\bmu, \ident)$---that is, from a Normal distribution with mean $\bmu$ and identity covariance matrix $\ident$.
    \item Given $\boldz$, draw each token $x_t \in \{1, \ldots, V\}$ in $x$ independently from a Categorical distribution with $\softmax(\boldW \boldz)$  as its parameters.
\end{enumerate}
Note that the above model closely resembles the shallow model of Section~\ref{sec:catlatmodel}, except that instead of using the latent variable $z$ to index into a set of Categorical distribution parameters, we let the parameters of a Categorical distribution be a function of the latent $\boldz$. We can thus view the latent variable $\boldz$ as a lower dimensional representation of our original sentence $x$. Letting $\pi = \{ \boldW\}$ and $\theta = \{\bmu \} \cup \pi$, we have the joint density
\begin{align} \label{eq:realnbdensity}
 p(x, \boldz ; \ \theta) = p(\boldz; \ \bmu) \times p(x \given \boldz \param \boldW) &= \mcN(\boldz; \bmu, \ident) \times  \prod_{t=1}^T p(x_t \given \boldz \param \ \boldW) \nonumber \\
 &= \mcN(\boldz; \bmu, \ident) \times  \prod_{t=1}^T \softmax(\boldW \boldz)_{x_t}.
 \end{align}
We show a corresponding graphical model in Figure~\ref{fig:realnbgm}. Note that the dependence structure in Figure~\ref{fig:realnbgm} is identical to that depicted in Figure~\ref{fig:nbgm}; we have, however, changed the type of latent variable (from discrete to continuous) and the parameterizations of the corresponding distributions.

\paragraph{Making the Model ``Deep''}
As in Section~\ref{sec:catlatmodel}, we may replace the Naive Bayes distribution over tokens with one parameterized by an RNN. We thus have the generative process:
\begin{enumerate}
    \item Draw latent variable $\boldz$ from $\mcN(\bmu, \ident)$.
    \item Given $\boldz$, draw each token $x_t \in \{1, \ldots, V\}$ in $x$ from a \textit{conditional} RNN, $\CRNNLM(x_{1:T} \param \pi, \boldz)$.
\end{enumerate}
We use the notation $\CRNNLM(x_{1:T} \param \pi, \boldz)$ to refer to the distribution over sentences induced by conditioning an ordinary $\RNNLM$ on some vector $\boldz$, by concatenating $\boldz$ onto the RNN's input at each time-step. In particular, we define
\begin{align} \label{eq:crnnlm}
    \CRNNLM(x_{1:T} \param \pi, \boldz) = \prod_{t=1}^T \softmax(\boldW \boldh_t)_{x_t},
\end{align}
where
\begin{align} \label{eq:crnn}
    \boldh_t = \RNN(\boldh_{t-1}, [\boldx_{t-1}; \boldz] \param \pi).
\end{align}
Compare Equations \eqref{eq:crnnlm} and \eqref{eq:crnn} with Equations \eqref{eq:rnnlm} and \eqref{eq:rnn} in Section~\ref{sec:intro}.

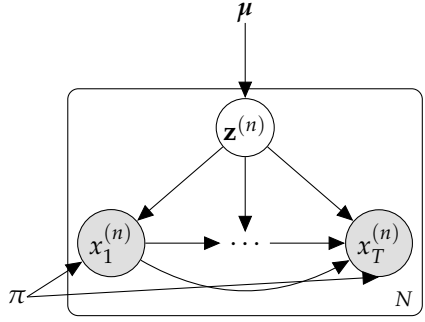
\begin{figure}
\centering
\begin{tikzpicture}
\node (dots) {$\ldots$};%
 \node[obs, left=1cm of dots] (x1) {$x_1^{(n)}$};%
 \node[obs, right=1cm of dots] (xT) {$x_T^{(n)}$};%
 \node[latent, above=of dots] (z) {$\boldz^{(n)}$}; %
 \node[const, above=of z] (mu) {$\bmu$};
 \node[const, below left=0.3cm and 0.8cm of x1] (pi) {$\pi$};
 
 \plate {plate1} {(dots)(x1)(xT)(z)} {$N$}; %
 \edge {z} {dots};
 \edge {z} {x1};
 \edge {z} {xT};
 \edge {mu} {z};
 \edge {x1} {dots};
  \edge {dots} {xT};
 \edge {pi.east} {x1,xT.south};
 
 \draw[->] 
 (x1) edge[bend right] node [right] {} (xT);
 
\end{tikzpicture}
 \caption{Continuous latent variable model with tokens generated by an RNN. The $\boldz^{(n)}$ have a normal distribution $\mcN(\bmu, \ident)$, and the $x^{(n)}$ have a $\CRNNLM(x^{(n)}_{1:T} \param \pi, \boldz^{(n)})$ distribution, where $\pi$ contains the parameters of the $\CRNNLM$. Note that the dependence structure is identical to that in Figure~\ref{fig:catrnngm}; the only difference is the type of latent variable and the parameterizations.}
 \label{fig:realrnngm}
\end{figure}

Then, letting $\pi$ contain the parameters of the $\CRNNLM$, we may write the joint density as
\begin{align} \label{eq:realrnndensity}
 p(x, \boldz ; \ \theta) = p(\boldz; \ \bmu) \times p(x \given \boldz \param \boldW) &= \mcN(\boldz; \bmu, \ident) \times \CRNNLM(x_{1:T} \param \pi, \boldz),
 \end{align}
 and we show the corresponding graphical model in Figure~\ref{fig:realrnngm}. As before, the dependence structure mirrors that in Figure~\ref{fig:catrnngm} exactly.

\paragraph{Inference}
Again using Bayes's rule, the posterior under continuous latent variable models is given by
\begin{align*}
    p(\boldz \given x \param \theta) = \dfrac{p(x, \boldz \param \theta)}{p(x \param \theta)} = \dfrac{p(\boldz \param \theta) p(x \given \boldz \param \theta)}{\int p(\boldz \param \theta) p(x \given \boldz \param \theta) \, \,d\boldz}.
\end{align*}
Unlike with discrete latent variables, however, calculating the denominator above will in general be intractable. Indeed, using deep models will generally prevent us from exactly calculating the denominator above even in relatively simple cases. This concern motivates many of the methods we discuss in later sections.

\subsection{A Structured, Discrete Latent Variable Model}
\label{sec:structlatmodel}
Finally, we will consider a model with multiple interrelated discrete latent variables per data-point. In particular, we will consider the Hidden Markov Model (HMM)~\citep{rabiner1989tutorial}, which models a sentence by assuming there is a discrete latent variable responsible for generating each word in the sentence, and that each of these latent variables depends only on the latent variable responsible for generating the previous word. Concretely, an HMM assumes that a sentence is generated according to the following process:

First, begin with variable $z_0$, which is always equal to the special ``start" state $0$ (i.e. $z_0 = 0$). Then, for $t = 1, \ldots, T$
\begin{enumerate}
    \item Draw latent variable $z_t \in \{1, \ldots, K\}$ from a Categorical distribution with parameter $\mu_{z_{t-1}} \in \Delta^{K-1}$.
    \item Draw observed token $x_t \in \{1, \ldots, V\}$ from a Categorical distribution with parameter $\pi_{z_{t}} \in \Delta^{V-1}$.
\end{enumerate}

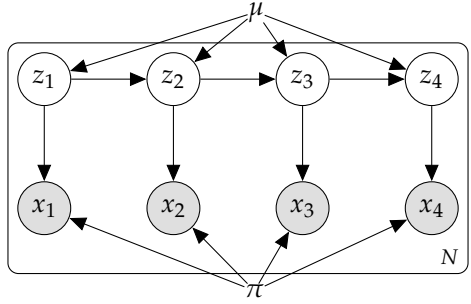
\begin{figure}
\centering
\begin{tikzpicture}
 \node[obs] (x1) {$x_1$};
 
 \node[obs, right=1cm of x1] (x2) {$x_2$};%
 \node[obs, right=1cm of x2] (x3) {$x_3$};%
 \node[obs, right=1cm of x3] (x4) {$x_4$};%
 \node[latent, above=of x1] (z1) {$z_1$}; %
 \node[latent, above=of x2] (z2) {$z_2$}; %
 \node[latent, above=of x3] (z3) {$z_3$}; %
 \node[latent, above=of x4] (z4) {$z_4$}; %
  \node[const, above left=0.5cm and 0.25cm of z3] (mu) {$\mu$};
  \node[const, below left=0.75cm and 0.25cm of x3] (theta) {$\pi$};

 

 \edge {mu} {z1,z2,z3,z4};
 \edge {z1} {x1,z2};
 \edge {z2} {x2,z3};
 \edge {z3} {x3,z4};
 \edge {z4} {x4};
 \edge {theta} {x1,x2,x3,x4};

 \plate {plate1} {(x1)(x2)(x3)(x4)(z1)(z2)(z3)(z4)} {$N$}; %
 \end{tikzpicture}
 \caption{HMM graphical model for a sequence of length $T=4$.}
 \label{fig:hmm}
\end{figure}

The above generative process gives rise to the following joint distribution over the sentence $x = x_1, \ldots, x_T$ and latent variables $z = z_1, \ldots, z_T$:
\begin{alignat}{2} \label{eq:hmm}
    p(x, z; \ \theta) &= p(z_1, \ldots, z_T \param \mu) &&\times p(x_1, \ldots, x_T \given z_1, \ldots, z_T \param \pi) \nonumber \\
    &= \prod_{t=1}^T p(z_t \given z_{t-1} \param \mu_{z_{t-1}}) &&\times \prod_{t=1}^T p(x_t \given z_{t} \param \pi_{z_{t}}) \nonumber \\ 
    &= \prod_{t=1}^T \mu_{z_{t-1}, z_t} &&\times \prod_{t=1}^T \pi_{z_{t}, x_t}
\end{alignat}
We show the corresponding graphical model in Figure~\ref{fig:hmm}.

It is important to note that the second equality above makes some significant independence assumptions: it assumes that the probability of $z_t$ depends only on $z_{t-1}$ (and not on $z_{<t-1}$ or $x_{<t}$), and it assumes that the probability of $x_t$ depends only on $z_t$ (and not on $z_{<t}$ or $x_{<t}$). These "Markov" (i.e., independence) assumptions are what give the Hidden Markov Model its name. We also note that we have referred to an HMM as a ``structured'' latent variable model because the latent sequence $z = z_1, \ldots, z_T$ is structured in the sense that it contains multiple components that are interdependent, as governed by $\mu$.

\paragraph{Making the Model ``Deep''}
We can create a ``deep'' HMM (c.f., \citet{tran2016,Johnson2016}) by viewing the $\mu_z$ and $\pi_z$ categorical parameters as being parameterizable in their own right, and parameterizing them with neural network components. 
For example, we might parameterize an HMM's categorical distributions as follows:
\begin{align*}
p(z_t \given z_{t-1}) = \softmax(\MLP(z_{t-1} \param \mu))  \qquad \qquad
p(x_t \given z_{t}) = \softmax(\MLP(z_{t-1} \param \pi)),
\end{align*}
where $\mu$ and $\pi$ now refer to the parameters of the corresponding MLPs. Note that the graphical model in Figure~\ref{fig:hmm} remains correct for the deep HMM; indeed, graphical models only show the dependency structure of the model (which has not been changed), and not the particular parameterization chosen. We also note that a deep parameterization may allow us to use fewer parameters as $K$ and $V$ get large. In particular, a standard HMM requires $O(K^2 + KV)$ parameters, whereas if the $\MLP$s above have $d$ hidden units we require only $O(Kd + Vd + d^2)$ parameters.

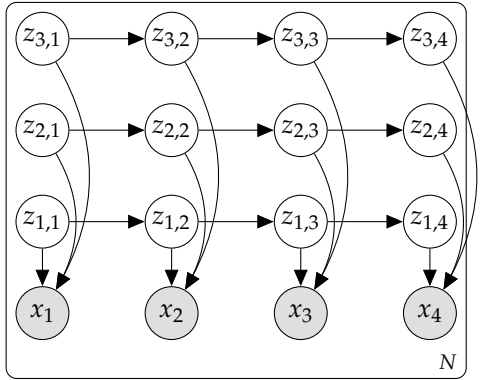
\begin{figure}
\centering
\begin{tikzpicture}
 \node[obs] (x1) {$x_1$};
  \node[obs, right=1cm of x1] (x2) {$x_2$};%
 \node[obs, right=1cm of x2] (x3) {$x_3$};%
 \node[obs, right=1cm of x3] (x4) {$x_4$};%

 \node[latent, above=5mm of x1] (z1) {$z_{1,1}$}; %
 \node[latent, above=5mm of x2] (z2) {$z_{1,2}$}; %
 \node[latent, above=5mm of x3] (z3) {$z_{1,3}$}; %
 \node[latent, above=5mm of x4] (z4) {$z_{1,4}$}; %
  \node[latent, above=5mm of z1] (k1) {$z_{2,1}$}; %
 \node[latent, above=5mm of z2] (k2) {$z_{2,2}$}; %
 \node[latent, above=5mm of z3] (k3) {$z_{2,3}$}; %
 \node[latent, above=5mm of z4] (k4) {$z_{2,4}$}; %
   \node[latent, above=5mm of k1] (j1) {$z_{3,1}$}; %
 \node[latent, above=5mm of k2] (j2) {$z_{3,2}$}; %
 \node[latent, above=5mm of k3] (j3) {$z_{3,3}$}; %
 \node[latent, above=5mm of k4] (j4) {$z_{3,4}$}; %

 

 \edge {z1} {x1,z2};
 \edge {z2} {x2,z3};
 \edge {z3} {x3,z4};
 \edge {z4} {x4};
  \edge {k1}{k2};
 \edge {k2} {k3};
 \edge {k3} {k4};
  \edge {j1}{j2};
 \edge {j2} {j3};
 \edge {j3} {j4};
 \draw[->] (k4) to[bend left=30] (x4);
 \draw[->] (k3) to[bend left=30] (x3);
 \draw[->] (k2) to[bend left=30] (x2);
 \draw[->] (k1) to[bend left=30] (x1);
  \draw[->] (j4) to[bend left=30] (x4);
 \draw[->] (j3) to[bend left=30] (x3);
 \draw[->] (j2) to[bend left=30] (x2);
 \draw[->] (j1) to[bend left=30] (x1);

 \plate {plate1} {(x1)(x2)(x3)(x4)(z1)(z2)(z3)(z4)(k1)(k2)(k3)(k4)(j1)(j2)(j3)(j4)} {$N$}; %
 \end{tikzpicture}
 \caption{Factorial HMM graphical model for a sequence of length $T=4$.}
 \label{fig:fhmm}
\end{figure}

\paragraph{Inference}
For structured, discrete latent variable models we have
\begin{align*}
    p(z \given x \param \theta) = \dfrac{p(x, z \param \theta)}{p(x \param \theta)} = \dfrac{p(z \param \theta) p(x \given z \param \theta)}{\sum_{z'} p(z = z' \param \theta) p(x \given z' \param \theta)},
\end{align*}
where we index all possible latent structures (e.g., all sequence of discrete latent variables of length $T$) with $z'$. It is possible to compute this sum over $z'$ structures with a dynamic program (e.g., the forward or backward algorithms~\citep{rabiner1989tutorial}), and for certain models, like HMMs, it will be tractable to do so. For other, more complicated structured latent variable models the dynamic program will be intractable to compute, necessitating the use of approximate inference methods. For instance, for the Factorial HMM (FHMM)~\citep{fhmm1996} depicted in Figure~\ref{fig:fhmm}, which generates each word $x_t$ by conditioning on the current state of \textit{several} independent first-order Markov chains, calculation of the denominator above will be exponential in depth and therefore intractable even with a dynamic program.

\section{Motivation and Examples}
\label{sec:motivation}
We now motivate the archetypal models we have introduced in the previous section by providing some examples of where similar models have been used in the literature. In general, we tend to be interested in latent variable models for any of the following interrelated reasons:
\begin{itemize}
\item[(a)] we have no or only partial supervision;
\item[(b)] we want to make our model more interpretable through $z$;
\item[(c)] we wish to model a multimodal distribution over $x$; 
\item[(d)] we would like to use latent variables to control our predictions and generations;
\item[(e)] we are interested in learning some underlying structure or representation;
\item[(f)] we want to better model the observed data;
\end{itemize}
and we will see that many of these apply to the models we will be discussing.

\subsection{Examples of Discrete Latent Variable Models}
As noted in Section~\ref{sec:models}, it is common to interpret discrete latent variable models as inducing a clustering over data. We now discuss several prominent NLP examples where discrete latent variable models can be interpreted in this way.

\paragraph{Document Clustering} The paradigmatic example application of categorical latent variable models to text is document clustering (i.e., unsupervised document classification). In this setup we are given a set of $N$ documents $\{x^{(n)}\}_{n=1}^N$, which we would like to partition into $K$ clusters such that intra-cluster documents are maximally similar. This clustering can be useful for retrieving or recommending unlabeled documents similar to some query document. Seen from the generative modeling perspective described in Section~\ref{sec:catlatmodel}, we view each document $x^{(n)}$ as being generated by its corresponding latent cluster index, $z^{(n)} \in \{1, \ldots, K\}$, which gives us a model of $p(x^{(n)}, z^{(n)} \param \theta)$. Since we are ultimately interested in obtaining the label (i.e., cluster index) of a document, however, we would then form the posterior over labels $p(z^{(n)} \given x^{(n)} \param \theta) = \frac{p(x^{(n)}, z^{(n)} \param \theta)}{\sum_{k=1}^K p(x^{(n)}, z^{(n)}=k \param \theta)}$ in order to determine the likely label for $x^{(n)}$.\footnote{Note that if we are just interested in finding the most likely label for $x^{(n)}$ we can simply evaluate $\argmax_k p(x^{(n)}, z^{(n)}=k \param \theta)$, since $\argmax_k p(x^{(n)}, z^{(n)}=k \param \theta) = \argmax_k \frac{p(x^{(n)}, z^{(n)} = k \param \theta)}{\sum_{k'=1}^K p(x^{(n)}, z^{(n)}=k' \param \theta)}$.}

Work on document clustering goes back decades (see \citet{willett1988recent} and \citet{aggarwal2012survey} for surveys), and many authors take the latent variable perspective described above. In terms of incorporating deep models into document clustering, it is possible to make use of neural components (such as RNNs) in modeling the generation of words in each document, as described in Section~\ref{sec:catlatmodel}. It has been more common recently, however, to attempt to cluster real-valued vector-representations (i.e., embeddings) of documents. For instance, it is common to pre-compute document or paragraph embeddings with an unsupervised objective and then cluster these embeddings with K-Means~\citep{macqueen1967some}; see ~\citet{le2014distributed} and \citet{xu2015short} for different approaches to obtaining these embeddings before clustering. \citet{xie2016unsupervised} continue to update document embeddings as they cluster, using an auxiliary loss derived from confidently clustered examples.  Whereas none of these document-embedding based approaches are presented explicitly within the generative modeling framework above, they can all be viewed as identifying parameterized distributions over real vectors (e.g., Gaussians with particular means and covariances) with each cluster index, which in turn generate the \textit{embeddings} of each document (rather than the words of each document). This sort of generative model is then (at least in principle) amenable to forming posteriors over cluster assignments, as above.

\paragraph{Mixtures of Experts}
Consider a supervised image captioning scenario, where we wish to produce a textual caption $x$ in response to an image $c$, and we are given a training set of $N$ aligned image-caption pairs $\{(c^{(n)}, x^{(n)})\}_{n=1}^N$. When training a model to generate captions similar to those in the training set, it may be useful to posit that there are in fact several ``expert'' captioning models represented in the training data, each of which is capable of generating a slightly different caption for the same image $c$. For instance, experts might differ in terms of the kind of language used in the caption (e.g., active vs. passive) or even in terms of what part of the image they focus on (e.g., foreground vs. whatever the human in the image is doing). Of course, we generally don't know beforehand which expert is responsible for generating which caption in the training data, but we can aim to capture the variance in captioning induced by these posited experts by identifying each value of a categorical latent variable $z$ with a different posited expert, and assuming that a caption $x^{(n)}$ is generated by first choosing one of these $K$ experts, which in turn generates $x^{(n)}$ conditioned on $c^{(n)}$. This type of model is known as a Mixture of Experts (MoE) model~\citep{jacobs1991adaptive}, and it gives rise to a joint distribution over $x^{(n)}$ and $z^{(n)}$, only this time we will also condition on $c^{(n)}$: $p(x^{(n)}, z^{(n)} \given c^{(n)} \param \theta)$. As in the previous example, the discrete latents $z^{(n)}$ induce a clustering, where examples are clustered by which expert generated them.

MoE models are widely used, and they are particularly suited to ensembling scenarios, where we wish to ensemble several different prediction models, which may have different areas of expertise. \citet{garmash2016ensemble} use an MoE to ensemble several expert neural machine translation models, and \citet{lee2016stochastic} use a related approach, called ``diverse ensembling,'' in training a neural image captioning system, showing that their model is able to generate more diverse captions as a result; \citet{he2018moe} find that MoE also lead to more diverse responses in machine translation, and \citet{gehrmann2018end} use a similar technique for generating descriptions of restaurant databases. There has also been work in text-generation that uses an MoE model per \textit{token}, rather than per sentence~\citep{yin2016neural,le2016lstm,yang2018breaking}.\footnote{Although these models have multiple discrete latents per data point, which suggests that we should perhaps consider them to be structured latent variable models, we will consider them unstructured since the interdependence between token-level latents is not made explicit in the probability model; correlations between these latents, however, are undoubtedly modeled to some extent by the associated RNNs.} See also ~\citet{eigen2013learning} and \citet{shazeer2017outrageously}, who use MoE \textit{layers} in constructing neural network architectures, and in the latter case for the purpose of language modeling.

Note that in the case of MoE models, we are interested in using a latent variable model not just because the training examples are not labeled with which experts generated them (reason (a) above), but also for reasons (c) and (d). That is, we attempt to capture different modes in the caption distribution (perhaps corresponding to different styles of caption, for example) by identifying a latent variable with each of these experts or modes. Similarly, we might hope to control the style of the caption by restricting our captioning system to one expert. When it comes to document clustering, on the other hand, we are interested in latent variables primarily because we don't have labeled clusters for our documents (reason (a)). Indeed, since we are interested primarily in the \textit{posterior} over $z^{(n)}$, rather than the joint distribution, reasons (c) and (d) are not applicable.

\subsection{Examples of Continuous Latent Variable Models}
Continuous latent variable models can often be interpreted as performing a dimensionality reduction of data, by associating a low-dimensional vector in $\reals^d$ with a data point. One of the advantages of this sort of dimensionality reduction is that it becomes easier to have a finer-grained notion of similarity between data points, by calculating their distance, for example, in the dimensionally reduced space. We now discuss two prominent examples.

\paragraph{Topic Models} Topic Models~\citep{blei2003latent} are an enormously influential family of latent variable models, which posit a set of $K$ latent ``topics,'' realized as distributions over words, which in turn generate each document $x^{(n)}$ in a collection. Each document $x^{(n)}$ in the collection moreover has a latent distribution $\boldz^{(n)} \in \Delta^{K-1}$ over these $K$ topics, governing how often it chooses to discuss a particular topic. Note that the per-document distribution over topics $\boldz^{(n)}$ in a topic model plays a similar role to the per-sentence vector $\boldz^{(n)}$ in the shallow model of Section~\ref{sec:reallatmodel}. Thus, in both cases we might, for instance, determine how similar two data points are by measuring the distance between their corresponding $\boldz^{(n)}$'s. 

\paragraph{Sentence Generation} Whereas, with the exception of Topic Models, NLP has historically preferred discrete latent variables, the tendency of deep models to deal with continuous \textit{embeddings} of objects has spurred interest in continuous latent variable models for NLP. For example, it has recently become quite popular to view a sentence as being generated by a latent vector in $\reals^d$, rather than by a latent label in $\{1, \ldots, K\}$ as in the previous subsection. Thus, \citet{Bowman2016} develop a latent vector model of sentence generation, where sentences are generated with a $\CRNNLM$, in a very similar way to that presented in Section~\ref{sec:reallatmodel}. This model and its extensions~\citep{Yang2017,hu2017toward,Kim2018}, represent the dominant approach to neural latent variable modeling of sentence generation at this time. 

As suggested above, viewing sentences as being generated by real vectors gives us a fine-grained notion of similarity between sentences, namely, the distance between the corresponding latent representations. The desire to compute similarities between sentences has also motivated extensive work on obtaining sentence embeddings~\citep{le2014distributed,kiros2015skip,joulin2016bag,conneau2017supervised,peters2018deep,pagliardini2018unsupervised,ruckle2018concatenated}, which can often be interpreted as continuous latent variables, though they need not be.

\subsection{Example of Structured, Discrete Latent Variable Models}
A major motivation for using structured, discrete latent variables in NLP is the desire to infer structured discrete objects, such as parse trees or sequences or segmentations, which we believe to be represented in the data, even when they are unannotated. 

\paragraph{Unsupervised Tagging and Parsing}
The simplest example application of structured, discrete latent variable modeling in NLP involves inducing the part-of-speech (POS) tags for a sentence. More formally, we are given a sentence $x = x_1, \ldots, x_T$, and we wish to arrive at a sequence $z = z_1, \ldots, z_T$ of POS tags, one for each word in the sentence. HMMs, as described in Section~\ref{sec:structlatmodel}, and their variants have historically been the dominant approach to arriving at a joint distribution $p(x_1, \ldots, x_T, z_1, \ldots, z_T \param \theta)$ over words and tags~\citep{brown1992class,merialdo1994tagging,smith2005contrastive,haghighi2006prototype,johnson2007doesn,toutanova2008bayesian,kirk2010,christodoulopoulos2010two,blunsom2011hierarchical,stratos2016unsupervised}. We may then predict the POS tags for a new sentence $x$ by calculating
\begin{align*}
    \argmax_{z_1, \ldots, z_T} p(z_1, \ldots, z_T \given x_1, \ldots, x_T \param \theta) = \argmax_{z_1, \ldots, z_T} p(x_1, \ldots, x_T, z_1, \ldots, z_T \param \theta).
\end{align*}

There is now a growing body of work involving deep parameterizations of structured discrete latent variable models for unsupervised parsing and tagging. For instance, \citet{tran2016} obtain good results on unsupervised POS tagging by parameterizing the transition and emission probabilities of an HMM with neural components, as described in Section~\ref{sec:structlatmodel}. In addition, just as recent approaches to neural document clustering have defined models that generate document \textit{embeddings} rather than the documents themselves, there has been recent work in neural, unsupervised POS tagging based on defining neural HMM-style models that emit word \textit{embeddings}, rather than words themselves~\citep{lin2015unsupervised,he2018}.

Unsupervised dependency parsing represents an additional, fairly simple example application of neural models with structured latent variables. In unsupervised dependency parsing we attempt to induce a sentence's dependency tree without any labeled training data. Much recent unsupervised dependency parsing is based on the DMV model of \citet{klein2004corpus} and its variants~\citep{headden2009improving,spitkovsky2010viterbi,spitkovsky2011unsupervised}, where there are multiple discrete latent variables per word, rather than one, as in POS tagging. In particular, the DMV model can be viewed as providing a joint distribution over the words $x_1, \ldots, x_T$ in a sentence and discrete latent variables $z_1, \ldots, z_{3T}$ representing each left and right dependent of each word, as well as a final empty left and right dependent for each word. As in the case of unsupervised tagging, we are primarily interested in predicting the most likely dependency tree given a sentence:
\begin{align*}
    \argmax_{z_1, \ldots, z_{3T}} p(z_1, \ldots, z_{3T} \given x_1, \ldots, x_T \param \theta) = \argmax_{z_1, \ldots, z_{3T}} p(x_1, \ldots, x_T, z_1, \ldots, z_{3T} \param \theta).
\end{align*}
Again, neural approaches divide between those that parameterize DMV-like models, which jointly generate the dependency tree and its words, with neural components~\citep{jiang2016unsupervised,cai2017crf,han2017dependency}, and those which define DMV-models that generate embeddings~\citep{he2018}.

\paragraph{Text Generation}
Structured latent variables are also widely used in text generation models and applications.
For instance, \citet{miao2016language} learn a summarization model by positing the following generative process for a document: we first sample a condensed, summary version of a document $z_{1:T_1}$ from an $\RNNLM(z_{1:T_1} \param \theta)$. Then, conditioned on $z_{1:T_1}$, we generate the full version of the document: $x_1, \ldots, x_{T_2}$ by sampling from a $\CRNNLM(x_{1:T_2} \param \theta, z_{1:T_1})$. This model gives us a joint distribution over documents $x^{(n)}_{1:T_2}$ and their corresponding summaries $z^{(n)}_{1:T_1}$. Moreover, a summarization system is then merely a system that infers the posterior over summaries $z^{(n)}_{1:T_1}$ given a document $x^{(n)}_{1:T_2}$.

Whereas in the model of \citet{miao2016language} a document $x$ is generated by conditioning solely on latent variables $z$, it is also common to posit generative models in which text $x$ is generated by conditioning both on some latent variables $z$ as well as on some additional contextual information $c$. In particular, $c$ might represent an image for an image captioning model, a sentence in a different language for a machine translation model, or a database for data-to-document generation model. In this setting, the latent variables $z$ would then represent some additional unobserved structure that, together with $c$, accounts for the observed text $x$. Some notable recent examples in this vein include viewing text $x$ as being generated by both $c$ as well as a shorter sequence of latent variables $z$~\citep{kaiser2018fast,roy2018theory}, a sequence of fertility latent variables $z$~\citep{gu2018nonautoregressive}, a sequence of iteratively refined sequences of latent variables $z$~\citep{lee2018deterministic}, or by a latent template or plan $z$~\citep{wiseman2018learning}. 

\section{Learning and Inference}\label{sec:learningin}
After defining a latent-variable model, we are typically interested in being able to do two related tasks: (1) we would like to be able to learn the parameters $\theta$ of the model, and (2) once trained, we would like to be able to perform \textit{inference} over the model. That is, we'd like to be able to compute the posterior distribution $p(z \given x \param \theta)$ (or approximations thereof) over the latent variables, given some data $x$. As we will see, these two tasks are intimately connected because learning often uses inference as a subroutine. On an intuitive level, this is because if we \textit{knew}, for instance, the value of $z$ given $x$, learning $\theta$ would be simple: we would simply maximize $p(x \given z_{\text{known}} \param \theta)$. Thus, as we will see, learning often involves alternately inferring likely $z$ values, and optimizing the model assuming these inferred $z$'s.

The dominant approach to learning latent variable models in a probabilistic setting is to maximize the  log marginal likelihood. This is equivalent to minimizing $\KL[p_\star(x) \Vert p(x \param \theta)]$, the KL-divergence between the
 true data distribution $p_\star(x)$ and the model distribution $p(x \param \theta)$, where
 the latent variable $z$ has been marginalized out. It is also possible to approximately minimize other divergences between $p_\star(x )$  and $p(x \param \theta)$, e.g. the Jensen-Shannon divergence or the Wasserstein distance.
In the context of deep latent variable models, such methods often utilize a separate model (\textit{discriminator}/\textit{critic}) which
learns to distinguish between samples from $p_\star(x)$ from samples from $p(x \param \theta)$. The generative model $\theta$ is trained ``adversarially" to fool the discriminator.
This gives rise to a family of models known as Generative Adversarial Networks (GANs) \citep{goodfellow2014generative}. While not the main
focus of the this tutorial, we review GANs
and their applications to text modeling in section~\ref{gan}.

\subsection{Directly Maximizing the Log Marginal Likelihood} \label{directll}
We begin with cases where the log marginal likelihood,  i.e.
\[ \log p(x \param \theta) = \log \sum_z p(x, z \param \theta) \]
is tractable to evaluate. (The sum should be replaced with an integral if $z$ is continuous). This is equivalent to assuming posterior inference is tractable, since 
\[ p(z \given x \param \theta) = \frac{p(x, z \param \theta)}{p(x \param \theta)}. \]

Calculating the log marginal likelihood is indeed tractable in some of the models that we have seen so far, such as categorical latent variable models where $K$ is not too big, or certain structured latent variable models (like HMMs) where dynamic programs allow us to efficiently sum over all the $z$ assignments. In such cases, maximum likelihood training of our parameters $\theta$ then corresponds to solving the following maximization problem:
\[\argmax_{\theta} \sum_{n=1}^N \log p(x^{(n)}; \, \theta), \]
where we have assumed $N$ examples in our training set.

In cases where $p(x, z ;\theta)$ is parameterized by a deep model, the above maximization problem
is not tractable to solve exactly. We will assume, however, that $ p(x, z ; \theta)$  is differentiable with respect to $\theta$. The main tool for optimizing such models, then, is gradient-based optimization. 
In particular, define the log marginal likelihood over the training set $x^{(1:N)} = [x_1, \dots, x_N]$  as
\[ L(\theta) = \log p(x^{(1:N)} \param \theta) = \sum_{n=1}^N \log p(x^{(n)} \param \theta) = \sum_{n=1}^N \log \sum_{z }p(x^{(n)}, z \param \theta).\]
The gradient is given by
\begin{align*}
\nabla_\theta L(\theta) &= \sum_{n=1}^N \frac{\nabla_\theta  \sum_{z} p(x^{(n)}, z\param \theta)}{p(x^{(n)} \param \theta)} \hspace{30mm} \text{(chain rule)}\\
&=\sum_{n=1}^N \sum_z \frac{ p(x^{(n)}, z \param \theta)}{p(x^{(n)} \param \theta))}\nabla_\theta \log p(x^{(n)}, z \param \theta)  \hspace{8mm} \text{(since $\nabla p(x,z) = p(x,z)\nabla \log p(x,z))$}\\
&= \sum_{n=1}^N \E_{p(z \given x^{(n)} \param \theta)} [\nabla_{\theta} \log p(x^{(n)}, z \param \theta) ]
\end{align*}
Note that the above gradient expression involves an expectation over the posterior $p(z \given x^{(n)} \param \theta)$, and is therefore an example of how inference is used as a subroutine in learning. With this expression for the gradient in hand, we may then learn by updating the parameters as
\[ \theta^{(i+1)} = \theta^{(i)} + \eta \nabla_\theta L(\theta^{(i)}),\]
where $\eta$ is the learning rate and $\theta^{(0)}$ is initialized randomly. In practice the gradient
is calculated over \textit{mini-batches} (i.e. random subsamples of the training set), and adaptive algorithms \citep{duchi2011,zeiler2012,Kingma2015} are often used.

\subsection{Expectation Maximization (EM) Algorithm}\label{em}
The Expectation Maximization (EM) algorithm \citep{dempster77em} is an iterative method for learning latent variable models with tractable posterior inference.  It maximizes a lower bound on the log marginal likelihood at each iteration.
Given randomly-initialized starting parameters $\theta^{(0)}$, the algorithm updates the parameters via the following alternating procedure:
\begin{enumerate}
\item E-step: Derive the posterior under current parameters $\theta^{(i)}$, i.e., $p(z \given x^{(n)} ; \, \theta^{(i)})$ for all $n = 1, \dots, N$.
\item M-step: Define the \emph{expected complete data likelihood} as 
\[ Q(\theta, \theta^{(i)}) = \sum_{n=1}^N \E_{p(z \given x^{(n)} ; \ \theta^{(i)})} [\log p(x^{(n)}, z ; \ \theta)] \]
Maximize this with respect to $\theta$, holding $\theta^{(i)}$ fixed
\[ \theta^{(i+1)} = \argmax_{\theta} Q(\theta, \theta^{(i)}) \]
\end{enumerate}
It can be shown that EM improves the log marginal likelihood at each iteration, i.e. 
\[\sum_{n=1}^N \log p(x^{(n)} ; \, \theta^{(i+1)}) \ge \sum_{n=1}^N \log p(x^{(n)} ; \, \theta^{(i)}). \]

As a simple example, let us apply the above recipe to the Naive Bayes model in section~\ref{naivebayes}, with $K=2$:
\begin{enumerate}
\item E-step: for each $n = 1, \dots, N$, calculate
\begin{align*} p(z \given x^{(n)} ; \ \theta^{(i)}) = \frac{p(x^{(n)}, z ; \ \theta^{(i)})}{\sum_{z' \in \{1, 2\}}p(x^{(n)} , z'; \ \theta^{(i)})} = \frac{(\mu_1^{(i)})^{\ind[z = 1]} (1-\mu_1^{(i)})^{\ind[z=2]}\prod_{t=1}^T \pi^{(i)}_{z, x_t^{(n)}}}{(\mu_1^{(i)}) \prod_{t=1}^{T} \pi^{(i)}_{1,x_t^{(n)}} + (1-\mu_1^{(i)}) \prod_{t=1}^{T} \pi^{(i)}_{2,x_t^{(n)}}}.
\end{align*}
Note that above we have written the prior over $z$ in terms of a single parameter $\mu_1$, since we must have $\mu_2 = 1 - \mu_1$.

\item M-step: The expected complete data likelihood is given by 
\begin{align*}
Q(\theta, \theta^{(i)})&= \sum_{n=1}^N \sum_{z} p(z \given x^{(n)} ; \ \theta^{(i)})( \log p(x^{(n)} \given z^{} ;\ \pi ) + \log p( z^{} ;\ \mu_1)) 
\end{align*}

To maximize the above with respect to $\mu_1$, we can differentiate and set the resulting expression to zero. Using the indicator notation $\ind[\cdot]$ to refer to a function that returns 1 if the condition in the bracket holds and 0 otherwise, we have   
\begin{align*}
\frac{\partial Q(\theta, \theta^{(i)})}{\partial \mu_1} &= \frac{\partial \sum_{n} \sum_{z} p(z \given x^{(n)} \param \theta^i) \log (\mu_1^{\ind[z=1]} (1-\mu_1)^ {\ind[z=2]})}{\partial \mu_1}   = 0 \\
&\implies \frac{\sum_{n=1}^N \sum_z p(z \given x^{(n)} ; \ \theta^{(i)}) \ind[z=1]}{\mu_1} = \frac{\sum_{n=1}^N \sum_z p(z \given x^{(n)} ; \ \theta^{(i)}) \ind[z=2]}{1-\mu_1} \\ 
&\implies \mu_1 (\sum_{n=1}^N \sum_z p(z \given x^{(n)} ; \ \theta^{(i)})) = \sum_{n=1}^N \sum_z p(z \given x^{(n)} ; \ \theta^{(i)}) \ind[z=1] \\
&\implies \mu_1^{(i+1)} = \frac{\sum_{n=1}^N \sum_z p(z \given x^{(n)} ; \ \theta^{(i)}) \ind[z=1]}{N} = \frac{\sum_{n=1}^N q^{(n)}_1}{N},
\end{align*}
where $q^{(n)}_1 = \sum_{z} p(z \given x^{(n)} \param \theta^{(i)}) \ind[z = 1] = \E_{p(z \given x^{(n)} \param \theta^{(i)})} \ind[z=1]$.
(We can verify that the above is indeed the maximum since $\frac{\partial^2 Q(\theta, \theta^{(i)})}{\partial \mu_1^2} < 0$). A similar derivation for $\pi_{z,v}$ yields
\begin{align*}
\pi_{z,v}^{(i+1)} = \frac{\sum_{n=1}^N q^{(n)}_z \sum_{t=1}^T \ind[x^{(n)}_t = v] / T}{\sum_{n=1}^N q^{(n)}_z}.
\end{align*}
Note that these updates are analogous to the maximum likelihood parameters of a Naive Bayes model in the supervised case, except
that the empirical counts  $\sum_{i=1}^N \ind[z^{(n)} = z]$ have been replaced with the \emph{expected counts} $\sum_{i=1}^N q_z^{(n)}$ under the posterior distribution. 
\end{enumerate}

Let us now consider using EM to learn the parameters of the RNN model introduced at the end of Section~\ref{sec:catlatmodel}. Here, the E-step is similar to the Naive Bayes model and follows straightforwardly from Bayes' rule:
\begin{align*} p(z \given x^{(n)} ; \ \theta^{(i)}) = \frac{p(x^{(n)}, z ; \ \theta^{(i)})}{\sum_{z' \in \{1, 2\}}p(x^{(n)} , z'; \ \theta^{(i)})} = \frac{(\mu_1^{(i)})^{\ind[z = 1]} (1-\mu_1^{(i)})^{\ind[z=2]}\prod_{t=1}^T p(x_t^{(n)} \given x_{<t}, z \param \theta^{(i)})}{\sum_{z'\in\{1,2\}} (\mu_1^{(i)})^{\ind[z' = 1]} (1-\mu_1^{(i)})^{\ind[z'=2]} \prod_{t=1}^{T} p(x_t^{(n)} \given x_{<t}, z' \param \theta^{(i)})}.
\end{align*}
Unlike with Naive Bayes, however, there is no closed-form update for the M step.\footnote{A closed-form update
exists for $\mu$ but not for the RNN parameters.} In this case, we can perform gradient-based optimization,
\[ \theta^{(i+1)} = \theta^{(i)} + \eta \nabla_\theta Q(\theta, \theta^{(i)}),\]
where $\eta$ is the learning rate and the gradient is given by
\[ \nabla_\theta Q(\theta, \theta^{(i)}) = \sum_{i=1}^N \E_{p(z \given x^{(n)} \param \theta^{(i)})} [\nabla_{\theta} \log p(x^{(n)}, z \param \theta)].\]
This variant of EM is sometimes referred to  as \emph{generalized expectation maximization}  \citep{dempster77em,neal1998,Murphy:2012:MLP:2380985}.
Note that the above expression is in fact the same as the gradient of the log marginal likelihood from the previous section, i.e. $\nabla_\theta Q(\theta, \theta^{(i)}) = \nabla_\theta L(\theta)$. Therefore, generalized
EM is equivalent to directly performing gradient ascent on the log marginal likelihood. This connection between 
EM and gradient ascent on the log marginal likelihood has been noted in the literature before \citep{salak2003,kirk2010,sutton2012introduction}, and is perhaps unsurprising given that backpropagation on the log marginal likelihood implicitly performs posterior inference \citep{eisner2016}. 

The connection between generalized EM and gradient ascent on the log marginal likelihood is particularly relevant to deep generative models, which will generally not admit an exact M-step. Practically speaking, we may avoid manually calculating the posteriors in the E-step and then taking gradient steps in $Q(\theta, \theta^{(i)})$, and instead take gradient steps directly on $\log p(x \param \theta)$; this is made especially convenient with automatic differentiation tools.\footnote{Note, though, that taking gradient steps to maximize $\log p(x \param \theta)$ is only equivalent to doing a \textit{single} gradient step during the $M$-step of generalized EM, and we are not guaranteed that the log marginal will increase monotonically as in standard EM.} For example, when training a deep
Hidden Markov Model (HMM) as described in Section~\ref{sec:structlatmodel},  we can use the forward (or backward) algorithm to calculate $\log p(x \param \theta)$, and call backpropagation
on the resulting value, instead of manually implementing the backward (or forward) algorithm to obtain the posteriors.\footnote{See \citet{kong2016segrnn}, \citet{yu2016online,yu2017noisy}, \citet{wiseman2018learning}, and \citet{kawakami2018segmental} for concrete examples of performing direct marginalization over latent variables in deep generative models.} 
Similarly, to train Probabilistic Context-Free Grammars (PCFG) with neural parameterizations of rule probabilities, we can run the inside algorithm to calculate
the log marginal likelihood and call backpropagation on the resulting value instead of manually implementing the outside algorithm.

Finally, we note that the EM algorithm can in general be been seen as performing coordinate ascent
on a lower bound on the log marginal likelihood~\citep{bishop2006prml}. This view (elaborated further in the next section)
will become useful when considering cases where posterior inference is \textit{intractable}, 
and will motivate \emph{variational inference}---a class of methods which uses approximate but tractable posteriors in place of the true posterior.

\subsection{Variational Inference}\label{vi}
So far we have considered models in which posterior inference (or equivalently, calculation of the log marginal likelihood) is tractable either via enumeration or dynamic programming. Now we consider cases in which posterior inference is intractable. Variational inference \citep{hinton1993,Jordan1999} is a technique for approximating an intractable posterior distribution
$p(z \given x \param \theta)$ with a tractable surrogate. In the context of learning the parameters of a latent variable model, variational inference can be used in optimizing a lower bound on the log marginal likelihood that involves only an \textit{approximate} posterior over latent variables, rather than the exact posteriors we have been considering until now.

We begin by defining
a set of distributions $\mathcal{Q}$, known as the \textit{variational family}, whose elements are 
distributions $q(z \param \lambda)$ parameterized by $\lambda$. That is, $\mcQ$ contains distributions over our latent variables $z$.
We will use $\mathcal{Q}$ to denote the entire variational family, and $q(z \param \lambda) \in \mathcal{Q}$
to refer to a particular variational distribution within the variational family, which is picked out by $\lambda$.
Let us assume that $z$ is continuous. We now derive a lower bound on the marginal log-likelihood $\log p(x \param \theta) = \log \int_z p(x, z \param \theta) dz$ that makes use of a $q(z \param \lambda)$ distribution.

We can lower bound the log marginal likelihood as follows:
\begin{align*}
\log p(x \param \theta) &= \int q(z \param \lambda) \log p(x \param \theta) \, dz && \text{(expectation of non-random quantity)} \\
&= \int q(z \param \lambda) \log \frac{p(x, z \param \theta)}{p(z \given x \param \theta)} \, dz && \text{(rewriting $p(x\param \theta)$, but see below)} \\
&= \int q(z \param \lambda) \log \left( \frac{p(x, z \param \theta)}{q(z \param \lambda)} \frac{q(z \param \lambda)}{p(z \given x \param \theta)} \right) \, dz && \text{(multiplying by 1)} \\
&= \int  q(z \param \lambda) \log \frac{p(x, z \param \theta)}{q(z \param \lambda)} \, dz + \int q(z \param \lambda)  \log \frac{q(z \param \lambda)}{p(z \given x \param \theta)} \, dz && \text{(distribute $\log$)} \\
&= \int q(z \param \lambda) \log \frac{p(x, z \param \theta)}{q(z \param \lambda)} \, dz + \KL[q(z \param \lambda)  \, \Vert \, p(z \given x \param \theta)]  && \text{(definition of KL divergence)} \\
&= \E_{q(z \param \lambda)} \log \frac{p(x, z \param \theta)}{q(z \param \lambda)} \quad \ + \KL[q(z \param \lambda)  \, \Vert \, p(z \given x \param \theta)] && \text{(definition of expectation)} \\
&= \ELBO(\theta, \lambda \param x) \qquad \qquad  \, + \KL[q(z \param \lambda)  \, \Vert \, p(z \given x \param \theta)] && \text{(definition of ELBO)} \\
&\geq \ELBO(\theta, \lambda \param x) && \text{(KL always non-negative)}
\end{align*}

The above derivation shows that $\log p(x \param \theta)$ is equal to a quantity called the \emph{evidence lower bound}, or ELBO, plus the KL divergence between $q(z \param \lambda)$ and the posterior distribution $p(z \given x \param \theta)$. Since the KL divergence is always non-negative, the ELBO is a lower-bound on  $\log p(x \param \theta)$, and it is this quantity that we attempt to maximize with variational inference. Before discussing the ELBO in more depth, we note that the above derivation requires that the support of the variational distribution lie within the support of the true posterior, i.e., $p(z \given x \param \theta) = 0 \implies q(z \param \lambda) = 0$ for all $z$.\footnote{Otherwise, the second equality would have a division by zero. In contrast, we can have $q(z \param \lambda) = 0$ and $p(z \given x \param \theta) > 0$ for some $z$, since the integral remains unchanged if we just integrate over the set $E = \{z :  q(z \param \lambda) >0\}.$} 

The form of the ELBO is worth looking at more closely. First, note
that it is a function of $\theta, \lambda$ (the data $x$ is fixed), and lower bounds the log marginal likelihood $\log p(x \param \theta)$ for any $\lambda$. The bound is tight
if the variational distribution equals the true posterior, i.e. $q(z \param \lambda ) = p(z \given x \param \theta)$  for all $z$ $\implies \log p(x \param \theta) = \ELBO(\theta, \lambda \param x)$. 
It is also immediately evident that 
\begin{align*}
\ELBO(\theta, \lambda \param x) = \log p(x \param \theta) - \KL[q(z \param \lambda)  \, \Vert \, p(z \given x \param \theta)].
\end{align*}

In some scenarios the model parameters $\theta$ are given (and thus fixed), and the researcher is tasked with finding the best variational approximation to the true posterior. Under this setup, $\log p(x \param \theta)$ is a constant and therefore maximizing the $\ELBO$ is equivalent to minimizing $\KL[q(z \param \lambda)  \, \Vert \, p(z \given x \param \theta)]$. 
However for our purposes we are also interested in learning the generative model parameters $\theta$.

\begin{figure}
    \centering
    \includegraphics[scale=0.25]{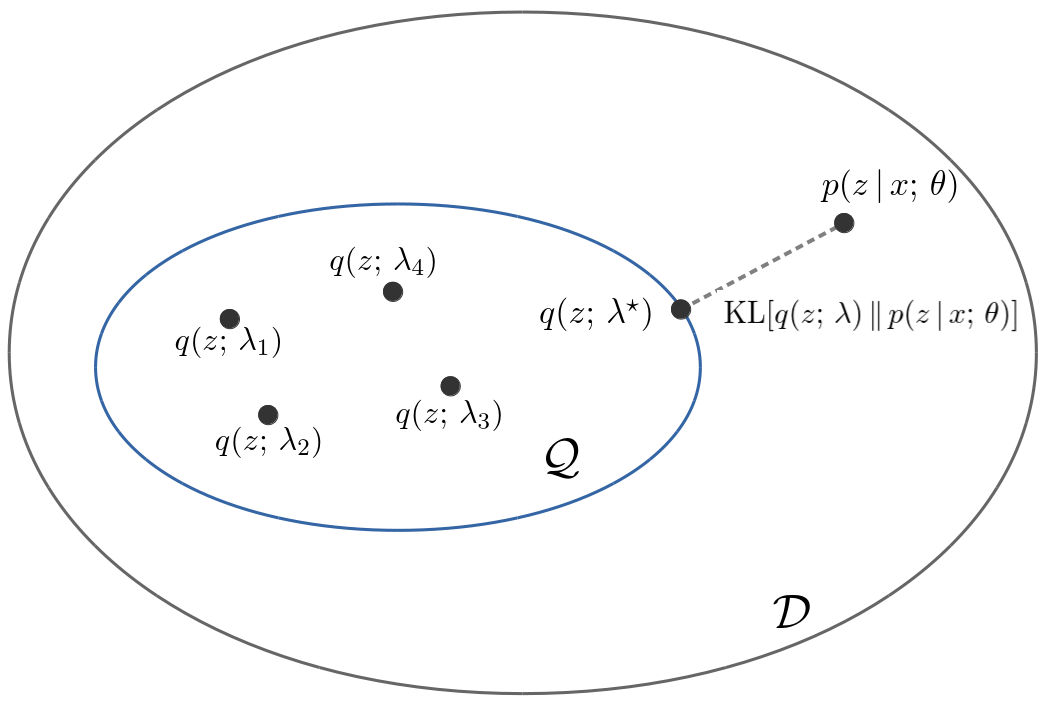}
    \caption{Illustration of variational inference. $\mathcal{D}$ represents all possible distributions
    over $z$, and $\mathcal{Q}$ represents the variational family with free parameters $\lambda$. Variational inference finds the distribution within $\mathcal{Q}$ that is closest to $p(z \given x \param \theta)$
    in the reverse KL sense, i.e. $\KL[q(z \param \theta) \Vert p(z \given x \param \theta)]$. In the above this  distribution is denoted as $q(z \param \lambda^\star)$.}
    \label{fig:vi}
\end{figure}

Letting $x^{(1:N)} = [x_1, \dots, x_N]$ be the training set, the ELBO over the entire dataset is given by the sum of individual ELBOs,
\[\ELBO(\theta, \lambda\param x^{(1:N)}) = \sum_{n=1}^N \ELBO(\theta, \lambda^{(n)} \param x^{(n)}) = \sum_{n=1}^N \E_{q(z \param \lambda^{(n)})}\Big[\log \frac{p(x^{(n)}, z \param \theta)}{q(z \param \lambda^{(n)})}\Big],  \]
where the variational parameters are given by $\lambda = [\lambda^{(1)}, \dots, \lambda^{(n)}]$ (i.e. we have $\lambda^{(n)}$ for each data point $x^{(n)}$).\footnote{There are many other possible parameterizations. For example, instead of nonparametric-style inference where  $\lambda$ is the union of local variational parameters $\lambda^{(n)}$, one could have a parameteric model with global parameters $\phi$ that is run over $x^{(n)}$ to produce the local variational parameters $\lambda^{(n)}$. This style of inference is called \emph{amortized variational inference} (see section~\ref{avi}).} Since $x^{(n)}$ are assumed to be drawn i.i.d, it is clear that the aggregate ELBO lower bounds the log likelihood of the training corpus
\[ \ELBO(\theta, \lambda\param x^{(1:N)}) \le \log p(x^{(1:N)} \param \theta).\]
It is this aggregate ELBO that we wish to maximize with respect to $\theta$ and $\lambda$ to train our model.

\subsubsection{Maximizing the ELBO}
One possible strategy for maximizing the aggregate ELBO is coordinate ascent, where we 
maximize the objective with respect to the $\lambda^{(n)}$'s keeping $\theta$ fixed, then maximize 
with respect to $\theta$ keeping the $\lambda^{(n)}$'s fixed. In particular, we arrive at the following:

\begin{enumerate}
\item Variational E-step: For each $n = 1, \dots, N$, maximize the ELBO for each $x^{(n)}$ holding $\theta^{(i)}$ fixed
\[ \lambda^{(n)} = \argmax_{\lambda} \ELBO(\theta^{(i)}, \lambda \param x^{(n)}) = \argmin_{\lambda}
\KL[q(z \param \lambda^{(n)})  \, \Vert \, p(z \given x^{(n)} \param \theta^{(i)})], \]
where the second equality holds since $\log p(x \param \theta^{(i)})$ is a constant with respect to the $\lambda^{(n)}$'s.
\item Variational M-step: Maximize the aggregated ELBO with respect to $\theta$ holding the $\lambda^{(n)}$'s fixed
\[ \theta^{(i+1)} = \argmax_{\theta} \sum_{n=1}^N \ELBO(\theta^{}, \lambda^{(n)} \param x^{(n)}) = 
\argmax_{\theta} \sum_{n=1}^N \E_{q(z \param \lambda^{(n)})}[\log p(x^{(n)}, z \param \theta)],  \]
where the second equality holds since the $\E_{q(z \param \lambda^{(n)})}[-\log q(z \param \lambda^{(n)})]$
portion of the ELBO is constant with respect to $\theta$.
\end{enumerate}
This style of training is also known as \emph{variational expectation maximization} \citep{neal1998}. In variational EM,
the E-step, which usually performs exact posterior inference, is instead replaced with variational inference which finds the best\footnote{In the ``reverse KL" sense, i.e. $\KL[q(z \param \lambda)  \, \Vert \, p(z \given x \param \theta)]$.} variational approximation to the true posterior. The E-step is illustrated in Figure~\ref{fig:vi}. The M-step maximizes the expected complete data likelihood where the expectation is taken with respect to the variational posterior.\footnote{Yet another variant of EM is \emph{Monte Carlo Expectation Maximization} \citep{wei1990mcem}, which obtains samples from the posterior in the E-step and maximizes the Monte Carlo estimate of the complete data likelihood in the M-step. In deep generative models sampling from the true posterior requires expensive procedures such as MCMC, although there has been some work on combining
variational inference with Hamiltonian Monte Carlo \citep{Salimans2015,hoffman2017}.}

If we consider the case where the variational family is flexible enough to include the true posterior,\footnote{That is, for all $x$ there exists $\lambda_x$ such that $q(z \param \lambda_x) = p(z \given x \param \theta)$ for all $z$.} then it is clear that the above reduces to the classic EM algorithm, since in the first step $\KL[q(z \param \lambda^{(n)})  \, \Vert \, p(z \given x^{(n)} \param \theta^{(i)})]$ is minimized when $q(z \param \lambda^{(n)})$ equals the true posterior.
Therefore, we can view EM as performing coordinate ascent on the ELBO where the variational family is arbitrarily
flexible. Of course, this case is uninteresting since we have assumed that exact posterior inference is
intractable. We are therefore interested in choosing a  variational family that is flexible enough
and at the same time allows for tractable optimization.  

In practice, performing coordinate ascent on the entire dataset is usually too expensive. The variational E-step can instead be performed over mini-batches.
As with generalized EM, the M-step can also be modified to perform gradient-based
optimization. It is also possible to perform the E-step only approximately, again using gradient-based optimization. This style of approach leads to a class of methods called \emph{stochastic variational inference (SVI)} \citep{Hoffman2013}. Concretely, 
for each $x^{(n)}$ in the mini-batch (of size $B$) 
we can randomly initialize $\lambda_0^{(n)}$ and 
perform gradient ascent on the ELBO with respect to $\lambda$ for $K$ steps,
\[ \lambda_{k}^{(n)} = \lambda_{k-1}^{(n)} + \eta \nabla_\lambda \ELBO(\theta, \lambda^{(n)}_k \param x^{(n)}), \,\,\,\,\,\,\,\,\, k = 1, \dots, K \]
Then the M-step, which updates $\theta$, proceeds with the variational parameters $\lambda_K^{(1)}, \dots,\lambda_K^{(B)}$ held fixed
\[ \theta^{(i+1)} = \theta^{(i)} + \eta \nabla_\theta \sum_{n=1}^B \E_{q(z \given \lambda_K^{(n)})}[\log p(x^{(n)}, z \param \theta^{(i)})]\]

In general, variational inference is a rich field of active research, and we have only covered a small portion of it in this section. For example, we have not covered
\emph{coordinate ascent variational inference}, which allows for closed-form updates in the E-step for conditionally conjugate models. We refer the reader to \cite{Wainwright2008}, \cite{Blei2017}, and \cite{Zhang2017} for further reading.

\section{Deep Inference}\label{sec:deepinf}
In the previous sections we have discussed two ways of performing inference---that is, of calculating posterior distributions. We have either calculated the exact posterior distribution $p(z \given x \param \theta)$ from its definition (i.e., $\frac{p(x, z \param \theta)}{p(x \param \theta)}$) in the case where doing so is tractable, or we have formed approximate posterior distributions $q(z \param \lambda)$ by optimizing variational parameters $\lambda$ so as to make $q(z \param \lambda)$ as close to the true posterior as possible. In this section we discuss a third alternative, whereby we simply train a neural network to \textit{predict} variational parameters $\lambda$, rather than arriving at $\lambda$ by optimizing the ELBO with respect to them. We refer to this latter strategy as ``deep inference.''

\subsection{Amortized Variational Inference and Variational Autoencoders}\label{avi}
Let us recall the variational expectation maximization algorithm from section~\ref{vi}. The variational E-step requires that we find the best variational parameters $\lambda^{(n)}$ for each $x^{(n)}$.
Even in mini-batch settings, this optimization procedure can be expensive, especially if a closed-form update is not available, which is typical in deep generative models. In such cases, one could rely on iterative methods to find approximately optimal variational parameters, as in SVI (see the previous section), but this may still be prohibitively expensive; indeed, each gradient calculation $\nabla_\lambda\ELBO(\theta, \lambda \param x^{(n)})$ requires backpropagating gradients through the generative model.

\begin{figure}
\center
    \begin{tikzpicture}
 \node[latent] (zl) {$z^{(n)}$};%
 \node[const, below=of zl] (lambda) {$\lambda^{(n)}$};
 \plate {plate1} {(zl)(lambda)} {$N$}; %
 \edge{lambda}{zl};
 
 \begin{scope}[xshift=5.3cm]
 \node[latent] (z) {$z^{(n)}$}; %
 \node[obs, below=of z] (x1) {$x^{(n)}$};%
 \node[const, right= of x1] (pi) {$\theta$};
 
 \plate {plate1} {(x1)(z)} {$N$}; %
 \edge {z} {x1};
 \edge {pi} {x1, z};
 \end{scope}
 \draw[dashed] (zl) --node [yshift=-0.35cm] {$\KL[q(z \param \lambda^{(n)} )\  ||\  p(z \given x^{(n)})]$} (z);
\end{tikzpicture}
\hspace{1cm}
    \begin{tikzpicture}
 \node[latent] (zl) {$z^{(n)}$};%
 \node[const, below=of zl] (xl) {$x^{(n)}$};%
 \node[const, above=of zl] (lambda) {$\phi$};
 \plate {plate1} {(zl)(xl)} {$N$}; %
 \edge{lambda}{zl};
 \edge{xl}{zl};
 
 \begin{scope}[xshift=5.7cm]
 \node[latent] (z) {$z^{(n)}$}; %
 \node[obs, below=of z] (x1) {$x^{(n)}$};%
 \node[const,  right= of x1] (pi) {$\theta$};
 
 \plate {plate1} {(x1)(z)} {$N$}; %
 \edge {z} {x1};
 \edge {pi} {x1, z};
 \end{scope}
 \draw[dashed] (zl) --node [yshift=-0.35cm] {$\KL[q(z \given x^{(n)} \param \phi) \, || \, p(z \given x^{(n)})]$} (z);
\end{tikzpicture}
\caption{(Left) Traditional variational inference uses variational parameters $\lambda^{(n)}$ for each data point $x^{(n)}$. (Right) Amortized variational inference
employs a global inference network $\phi$ that is run over the input $x^{(n)}$ to produce the local variational distributions.}
    \label{fig:avi}
\end{figure}
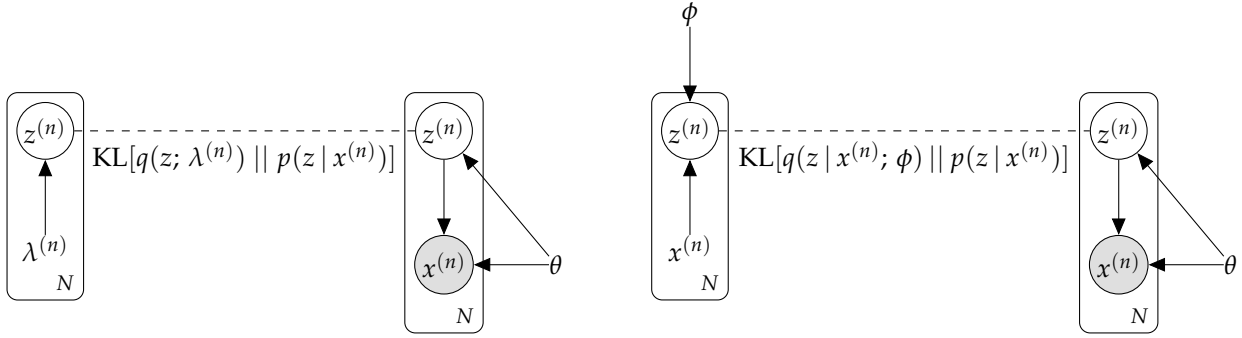

As an alternative, one could \emph{predict} the variational parameters 
by applying a trained neural network, called an \emph{inference network},\footnote{Also referred to a \textit{recognition network} or an \emph{encoder}.} to the input $x^{(n)}$ for which we would like to calculate an approximate posterior:
\[ \lambda^{(n)} = \enc(x^{(n)} \param \phi). \]
The inference network is trained (via gradient ascent) to perform variational inference for all the data points, i.e.
\[ \max_{\phi} \sum_{n=1}^{N} \ELBO(\theta, \enc(x^{(n)} \param \phi) \param x^{(n)}).\]
Importantly, the same encoder network (with parameters $\phi$) can be used for all $x^{(n)}$ we are interested in, and it is therefore unnecessary to optimize separate $\lambda^{(n)}$ for each $x^{(n)}$ we encounter. 
This style of inference is
also known as \emph{amortized variational inference} (AVI), as the task of performing  approximate posterior inference is \emph{amortized} across the entire dataset through the shared encoder. This is illustrated in is Figure~\ref{fig:avi}. AVI is usually much faster than both SVI and traditional VI, as one can simply run the inference network over $x^{(n)}$ to obtain the variational parameters, which should approximate the true posterior well if the inference network is sufficiently expressive and well-trained.

\emph{Variational autoencoders (VAEs)} \citep{Kingma2014,Rezende2014,Mnih2014} are a family of 
deep generative models where the variational parameters are predicted from a deep inference network
over the input, in the way just described. The \emph{autoencoder} part of the name stems from the fact that for generative models that factorize as  $p(x, z \param \theta) = p(x \given z \param \theta)p(z \param \theta)$, we can rearrange the ELBO as follows:
\begin{align} \label{eq:vaeelbo}
\ELBO(\theta, \phi \param x) &= \E_{q(z \given x \param \phi )} \Big [\log \frac{p(x, z \param \theta)}{q(z \param \phi)}\Big] \nonumber \\
&= \E_{q(z \given x \param \phi )} \Big[ \log \frac{p(x \given z \param \theta) p(z \param \theta)}{q(z \param \phi)}\Big] \nonumber \\ 
&= \E_{q(z \given x \param \phi )}[\log p(x \given z \param \theta)] - \KL[q(z \given x \param \phi)  \Vert  p(z \param \theta)].
\end{align}

Above, for brevity we have written $ \ELBO(\theta, \phi \param x)$ in place of $\ELBO(\theta, \enc(x \param \phi) \param x)$, and $q(z \given x \param \phi)$ in place of $q(z \param \enc(x \param \phi))$, and we will use this notation going forward. Note that the first term in the $\ELBO$ is the expected reconstruction likelihood of $x$ given the latent variables $z$, which is roughly equivalent to an autoencoding objective,
and the second term can be viewed as regularization term that pushes the variational distribution to be similar to the prior.

In the standard VAE setup, the inference network and the generative model are jointly trained by maximizing the ELBO with gradient ascent:
\begin{align*}
\theta^{(i+1)} &= \theta^{(i)} + \eta \nabla_{\theta} \ELBO(\theta^{(i)}, \phi^{(i)} \param x^{(n)}) \\
\phi^{(i+1)} &= \phi^{(i)} + \eta \nabla_{\phi} \ELBO(\theta^{(i)}, \phi^{(i)} \param x^{(n)}).
\end{align*}
The above updates are for a single data point, but in practice mini-batches are used. Note that unlike 
the coordinate ascent-style training from previous section, $\theta$ and $\phi$ are trained
together end-to-end. In the next subsection, we will illustrate VAE training in more depth using a simple text example. Table~\ref{tab:summary} summarizes the different optimization methods
we have encountered so far.

\subsection{Training a Text VAE}\label{app}
Let us revisit the model described in section~\ref{sec:reallatmodel} and see how it may be trained with a VAE. For simplicity, we will fix the mean and variance of the prior distribution over $\boldz$, and thereby arrive at the following generative process, as originally proposed by \citet{Bowman2016}:
\begin{enumerate}
    \item Sample $\boldz \sim \mathcal{N}(\mathbf{0}, \mathbf{I})$ with $\boldz \in \reals^d$.
    \item Sample $x \sim \CRNNLM(x \param \theta, \boldz)$. 
\end{enumerate}
We now define the variational family $\mathcal{Q}$ to contain Gaussian distributions with diagonal covariance matrices. That is, we define $q(\boldz \given \boldx \param \phi) = \mcN(\boldz \param \bmu, \mathrm{diag}(\bsigma^2))$, where
$[\boldsymbol{\mu}, \boldsymbol{\sigma^2} ] = \enc(\mathbf{x} \param \phi)$, and where $\bmu \in \reals^d$ and $\bsigma^2 \in \reals_{\geq 0}^d$. 
A popular parameterization for $\enc(x \param \phi)$ is 
\begin{align*}
    \boldh_{1:T} = \RNN(x), & & \boldh = \MLP(\boldh_T), & & \boldsymbol{\mu} = \mathbf{W}_1 \boldh + \boldb_1,  && \boldsymbol{\sigma}^2 = \exp(\mathbf{W}_2 \boldh + \boldb_2 ).
\end{align*} 
\subsubsection{Optimizing the ELBO and the Reparameterization Trick}
Given the above choice of variational family and approach to predicting $\lambda$, we may now attempt to maximize the ELBO with respect to $\theta$ and $\phi$. The overarching strategy is to rearrange some terms to express gradient of the expectation as an expectation of the gradient, which can be estimated with Monte Carlo
samples.

\begin{table}[]
    \centering
    \begin{tabular}{l l l }
    \toprule
        Method  & E-step & M-step \\
    \midrule
         Expectation Maximization & Exact Posterior: $q(z) = p(z \given x \param \theta)$ & Exact \\
         Log Marginal Likelihood & Exact Posterior: $q(z) = p(z \given x \param \theta)$ & Gradient \\
         Variational EM & VI: $q(z \param \lambda), \,\,\,\, \lambda = \argmax_{\lambda} \ELBO(\theta, \lambda \param x)$ & Exact/Gradient \\
         Stochastic Variational EM & Stochastic VI: $q(z \param \lambda), \,\,\,\, \lambda = \lambda + \eta \nabla_\lambda \ELBO(\theta, \lambda \param x)$ & Gradient\\
         Variational Autoencoder & Amortized VI: $q(z \param \lambda), \,\,\,\, \lambda = \enc(x \param \phi)$ & Gradient \\
         \bottomrule
    \end{tabular}
    \caption{Overview of the different optimization methods for training generative models. The ``Expectation" or ``Inference" step (E-step) correponds to performing posterior inference, i.e. minimizing $\KL[ q(z) \Vert p( z \given x \param \theta)]$. The ``Maximization" or ``Learning" step (M-step) corresponds to maximizing the complete data likelihood under the inferred posterior, i.e. $\E_{q(z)}[\log p(x, z \param \theta)]$.}
    \label{tab:summary}
\end{table}
The gradient of the ELBO with respect to $\theta$ is given by
\begin{align*}
    \nabla_\theta \ELBO(\theta,\phi \param x) &= \E_{q(\boldz \given x \param \phi)}[\nabla_\theta \log p(x, \boldz \param \theta)] \\ &= \E_{q(\boldz \given x \param \phi)} [\nabla_\theta \log p(x \given \boldz \param \theta)].
\end{align*}
The first equality holds because the distribution with which we are taking the expectation does not depend on $\theta$, so we can push the gradient inside the expectation.\footnote{Throughout this tutorial we will always assume that we can differentiate under the integral sign (i.e. swap the gradient/integral signs). This is valid under mild conditions, e.g. conditions which satisfy the hypotheses of the dominated convergence theorem.} The second equality holds because the prior (spherical Gaussian) 
does not depend on $\theta$.
The expectation in the gradient above is typically estimated with Monte Carlo samples, and one sample is often sufficient.

Let us now derive the gradient of the ELBO with respect to $\phi$ in the general case (i.e. for any generative model/variational distribution),
\begin{align*}
    \nabla_\phi \ELBO(\theta, \phi \param x) &= \nabla_\phi \E_{q(\boldz \given x \param \phi)} \Big[\log \frac{p(x, \boldz \param \theta)}{q(\boldz \given x \param \phi)} \Big] \\
    &= \nabla_\phi \E_{q(\boldz \given x \param \phi)} [\log p(x, \boldz \param \theta)] - \nabla_\phi \E_{q(\boldz \given x \param \phi)} [\log q(\boldz \given x \param \phi)].
\end{align*}
Unlike the case with $\theta$, we cannot simply push the gradient sign
inside the expectation since the distribution with which we are taking the expectation depends on $\phi$.

We derive the gradients of the two terms in the above expression separately. The first term involves the standard score function gradient estimator, \citep{Glynn1987,Williams1992,Fu2006}
\begin{align*}
\nabla_\phi \E_{q(\boldz \given x \param \phi)} [ \log p(x, \boldz \param \theta)] &= 
\nabla_\phi \int  \log p(x , \boldz \param \theta) q(\boldz \given x \param \phi) \,d\boldz \\
&= 
\int  \log p(x, \boldz \param \theta) \nabla_\phi  q(\boldz \given x \param \phi) \,d\boldz && \text{(differentiate under the integral sign)} \\
&= 
\int  \log p(x , \boldz \param \theta)   q(\boldz \given x \param \phi)\nabla_\phi  \log q(\boldz \given x \param \phi) \,d\boldz && \text{(since $\nabla q = q \nabla \log q $)} \\
&=\E_{q(\boldz \given x \param \phi)} [\log p(x, \boldz \param \theta)\nabla_\phi \log q(\boldz \given x \param \phi)].
\end{align*}
The second term is given by,
\begin{align*}
\nabla_\phi \E_{q(\boldz \given x \param \phi)} &[\log q(\boldz \given x \param \phi)] = \nabla_\phi \int  \log q(\boldz \given x \param \phi) q(\boldz \given x \param \phi) \,d\boldz \\
&=  \int \nabla_\phi \Big(\log q(\boldz \given x \param \phi) q(\boldz \given x \param \phi) \Big) \,d\boldz && \text{(differentiate under the integral sign)}  \\
&=  \int q(\boldz \given x \param \phi)\nabla_\phi \log q(\boldz \given x \param \phi)  + \log q(\boldz \given x \param \phi)  \nabla_\phi  q(\boldz \given x \param \phi)  \,d\boldz && \text{(product rule)}  \\
&=  \int \nabla_\phi q(\boldz \given x \param \phi)\, d\boldz  + \int \log q(\boldz \given x \param \phi)  \nabla_\phi  q(\boldz \given x \param \phi)  \,d\boldz && \text{(apply $\nabla q = q \nabla \log q $ to first term)}  \\
&=  0 + \int \log q(\boldz \given x \param \phi)  \nabla_\phi  q(\boldz \given x \param \phi)  \,d\boldz && \text{(since $\int \nabla q = \nabla \int q = \nabla 1 = 0$)}  \\
&= \int \log q(\boldz \given x \param \phi) q(\boldz \given x \param \phi)   \nabla_\phi   \log q(\boldz \given x \param \phi)  && \text{(apply $\nabla q = q \nabla \log q $ again)} \\
&= \E_{q(\boldz \given x \param \phi)} [\log q(\boldz \given x \param \phi)\nabla_\phi \log q(\boldz \given x \param \phi)]
\end{align*}
Putting it all together, we have
\begin{align*}
    \nabla_\phi \ELBO(\theta, \phi \param x) = \E_{q(\boldz \given x \param \phi)}\Big[ \log \frac{p(x, \boldz \param \theta)}{q(\boldz \given x \param \phi)} \nabla_\phi \log q(\boldz \given x \param \phi) \Big],
\end{align*}
which is just policy gradient-style reinforcement learning with reward given by
$\log \frac{p(x, \boldz \param \theta)}{q(\boldz \given x \param \phi)}$. The expectation can again be estimated with Monte Carlo samples. While the Monte Carlo gradient estimator is unbiased it will suffer from high variance, and in this case a single sample is often \emph{not} sufficient.

However, we can exploit our choice of the variational family (i.e. $q(\boldz \given x \param \phi) =\mcN(\boldz \param \bmu, \mathrm{diag}(\bsigma^2)$) to derive
another estimator. First, using the KL decomposition of the ELBO, we observe that
we can also express the gradient of the ELBO with respect to $\phi$ as
\[ \nabla_\phi \ELBO(\theta, \phi \param x) = \nabla_\phi \E_{q(\boldz \given x \param \phi)} [ \log p(x \given \boldz \param \theta)] - \nabla_\phi \KL[q(\boldz \given x \param \phi)  \Vert  p(\boldz)].\]

Beginning with the second term, the KL divergence between a diagonal Gaussian and the standard Gaussian has an analytic solution given by
\[\KL[q(\boldz \given x \param \phi) \Vert p(\boldz)]  = -\frac{1}{2}\sum_{j=1}^d (\log \sigma^2_j - \sigma^2_j - \mu_j^2 + 1), \]
and therefore $\nabla_\phi \KL[q(\boldz \given x \param \phi)  \Vert  p(\boldz)]$ is easy to calculate.
For the first term, notice that our variational family of Gaussian distributions is \emph{reparameterizable}~\citep{Kingma2014,Rezende2014,glasserman2013monte} in
the sense that we can obtain a sample from the variational posterior by sampling from a base noise distribution and applying a deterministic transformation, 
\begin{align*} 
\boldsymbol{\epsilon} \sim \mcN(\mathbf{0}, \mathbf{I}) &&
\boldz = \boldsymbol{\mu} + \boldsymbol{\sigma} \boldsymbol{\epsilon},
\end{align*}
where $\bmu$ and $\bsigma^2$ are as usual given by our encoder network. Observe that $\boldz$ remains distributed according to $\mcN(\boldz \param \bmu, \mathrm{diag}(\bsigma^2))$, but 
we may now express the gradient with respect to $\phi$ as
\begin{align*}
\nabla_\phi \E_{q(\boldz \given x \param \phi)} [ \log p(x \given \boldz \param \theta)] &=
\nabla_\phi  \E_{\boldsymbol{\epsilon} \sim \mathcal{N}(\mathbf{0}, \mathbf{I})}[\log p(x \given \boldsymbol{\mu} + \boldsymbol{\sigma} \boldsymbol{\epsilon} \param \theta)] \\
&=  \E_{\boldsymbol{\epsilon} \sim \mathcal{N}(\mathbf{0}, \mathbf{I})}[\nabla_\phi \log p(x \given \boldsymbol{\mu} + \boldsymbol{\sigma} \boldsymbol{\epsilon} \param \theta)], 
\end{align*}
because the expectation no longer depends on $\phi$. We can again approximate the expectation in the gradient above with a single sample.

This \emph{reparameterization trick} just discussed empirically yields much lower-variance gradient estimators and
has been instrumental in training deep generative models with VAEs. Intuitively, 
the reparameterized gradient estimator ``differentiates through" the generative model and therefore has more information than
the score function gradient estimator, which treats the generative model as a black-box
reward function. 
Unfortunately, in many cases the
variational posterior is not directly reparameterizable (e.g. if $z$ is discrete), and extending the
reparameterization trick to other families of distributions is an active research area (see section~\ref{reparam}).\footnote{The term \emph{reparameterization trick} is a slight misnomer, since  we can reparameterize (for example) a discrete distribution by applying a deterministic transformation to base Gumbel noise (section~\ref{reparam}). However we will not be able to train $\phi$ with gradient-based optimization in this case, since gradients will be zero almost everywhere. In VAE literature the term reparameterizable has come to colloquially  mean ``reparameterizable through differentiable transformations that provide nonzero gradients."}

\subsubsection{Posterior Collapse}\label{posteriorcollapse}
We now discuss an important issue that affects training Text VAEs in practice.
Recall that in the model introduced in Section~\ref{sec:reallatmodel} the likelihood model is allowed to fully condition on the entire history, through the RNN's state $\boldh_t$:
\begin{align*} 
\boldh_{t} &= \RNN(\boldh_{t-1}, [\boldx_{t-1}; \boldz]) \\
p(x_{t} \given x_{< t}, \boldz \param \theta) &= \softmax(\boldW \boldh_{t})_{x_t},
\end{align*}
where the above equations are copied from Equations~\eqref{eq:crnn} and~\eqref{eq:crnnlm}.
We might hope that in such a model $\boldz$ 
would capture global aspects of the sentence/document (e.g., the topic), while the RNN learns to model local variations 
(e.g. placement of function words to ensure grammatically).\footnote{There has also been prior work on separating topics and syntax in the context of non-neural generative models \citep{griffiths2004,boyd2008}.} 

However, \citet{Bowman2016} observe that  these types of models
experience \emph{posterior collapse}, whereby the likelihood (i.e., reconstruction)
model ignores the latent variable  and simply becomes a language model. That is, $x$ and $\boldz$ 
become independent. Indeed, looking at the ELBO in Equation~\eqref{eq:vaeelbo}, we see that if $x$ can be reconstructed \textit{without} $z$, the model is incentivized to make the variational posterior approximately equal to the prior---that is, to have
\[ \KL[q(\boldz \given x \param \phi) \Vert p(\boldz)] \approx 0\]
regardless of how expressively one parameterizes $q(\boldz \given x \param \phi)$.  More formally, \citet{Chen2017}
show that this phenomenon may be justified under the ``bits-back" argument: if the likelihood
model is rich enough to model the true data distribution $p_\star(x)$ without using any information
from $\boldz$, then the global optimum is obtained by setting 
\begin{align*}
    p(x \given \boldz \param \theta) &= p(x \param \theta) = p_\star(x) \\
    p(\boldz \given x \param \theta) &=  q(\boldz \given x \param \phi) =p(\boldz).
\end{align*}
Since any distribution $p(x)$ can be factorized as $p(x) = p(x_1)\prod_{t=2}^T p(x_t \given x_{<t})$, it is possible that a large enough RNN can model $p_\star(x)$ without
relying on $\boldz$. As such, to avoid posterior collapse, past work has made conditional independence assumptions and instead used multilayer perceptrons \citep{Miao2016,miao2017nvi} or convolutional networks \citep{Yang2017,Semeniuta2017,shen2018deconv} to parameterize the likelihood model.

One way to mitigate posterior collapse from an optimization perspective is 
to ``warm-up" the KL portion of the objective \citep{Bowman2016,Son2016}. In particular, 
one maximizes
\[ \E_{q(\boldz \given x \param \phi)}[\log p(x \given \boldz \param \phi)] - \beta  \KL[q(\boldz \given x \param \phi) \Vert p(\boldz)], \]
where $\beta$ is gradually increased from $0$ to $1$ over training.\footnote{$\beta$-VAEs are a class of models where the multiplier $\beta$ on the KL portion of the ELBO is not necessarily $1$ \citep{higgins2017}. These models can be interpreted as the Langragian of a constrained optimization problem
with respect to the original ELBO.
Models trained with $\beta > 1$ have been shown to learn more \emph{disentangled}
latent representations, and using VAE-like models to learn more disentangled/factorized representations is an active
research area \citep{Burgess2017,kim2018dist,Chen2018vae}.} 
Other strategies include adding auxiliary objectives which ensure that $\boldz$ is used \citep{Dieng2017,Goyal2017b,Wang2018}, randomly dropping out words during
decoding \citep{Bowman2016}, thresholding the KL function so that some bits are ``free" \citep{Kingma2016},
combining amortized and stochastic variational inference \citep{Kim2018},
using skip connections \citep{dieng2018}, or working
with distributions where the KL can be essentially a fixed hyperparameter (e.g. the von Mises--Fisher distribution with a fixed concentration parameter \citep{Guu2017,xu2018}).
\subsubsection{Evaluation}\label{eval}
As the ELBO always lower bounds the log marginal likelihood $\log p(x \param \theta)$, we can evaluate
the learned generative model by evaluating the ELBO across a held-out test set. 
We can also estimate the marginal likelihood with $K$ importance samples,

\[ p(x \param \theta) = \E_{q(z \given x \param \phi)}\Big[ \frac{p(x, z \param \theta)}{q(z \given x \param \phi)} \Big] \approx  \frac{1}{K} \sum_{k=1}^K \frac{p(x, z^{(k)} \param \theta)}{q(z^{(k)} \given x \param \phi)},\]
where the samples are from the approximate variational posterior.

Therefore the log marginal likelihood can be estimated by
\[ \log p(x \param \theta) = \log \E_{q(z \given x \param \phi)}\Big[ \frac{p(x, z \param \theta)}{q(z \given x \param \phi)} \Big] \approx \log   \frac{1}{K} \sum_{k=1}^K \frac{p(x, z^{(k)} \param \theta)}{q(z^{(k)} \given x \param \phi)}.\]
The above estimator is not unbiased but under mild conditions converges almost surely to
$\log p(x \param \theta)$ as $K \rightarrow \infty$.\footnote{If $\frac{p(x,z \param \theta)}{q(z \given x \param \phi)}$ is bounded then by the strong law of large numbers $\frac{1}{K} \sum_{k=1}^K \frac{p(x, z^{(k)} \param \theta)}{q(z^{(k)} \given x \param \phi)} \rightarrow p(x \param \theta)$ almost surely as $K \rightarrow \infty$.
Then the continuous mapping theorem implies $\log \frac{1}{K} \sum_{k=1}^K \frac{p(x, z^{(k)} \param \theta)}{q(z^{(k)} \given x \param \phi)} \rightarrow \log p(x \param \theta)$ almost surely as $K \rightarrow \infty$.} It is also possible to maximize the above quantity directly, leading to \emph{importance weighted autoencoders} \citep{Burda2015} (see section~\ref{importancesampling}).
\cite{Wu2017quant} further refine the estimate of the log marginal likelihood with annealed
importance sampling \citep{neal2001}.

The ELBO and the log marginal likelihood provide a quantitative estimate of the learned generative model, 
but from a representation learning perspective these metrics may not be useful.
Indeed, \citet{alemi2018} find that in many cases there is a family of models that achieve
the same ELBO but have different reconstruction/KL terms.\footnote{To be precise,
\cite{alemi2018} use the term \emph{rate} and \emph{distortion}. Distortion corresponds to the negative
reconstruction likelihood portion of the ELBO, and rate corresponds to the $\KL [q(z \given x \param \phi) \Vert m(z)]$,
where $m(z)$ is a \emph{variational marginal} distribution. Hence the rate is not necessarily
equal to the KL portion of the ELBO, i.e. $\KL [q(z \given x \param \phi) \Vert p(z)]$. See the paper
for more discussion around $m(z)$.} It is therefore often useful to report the reconstruction 
term ($\E_{q(z \given x \param \phi)}[\log p( x \given z \param \theta)]$) and the KL term ($\KL [q(z \given x \param \phi) \Vert p(z)]$) along with the ELBO \citep{Bowman2016,Gulrajani2017,Yang2017}. The mutual information between $x$ and $z$ is also estimable though it may be computationally expensive \citep{hoffman2016elbo}.\footnote{Some works directly optimize an approximation to the mutual information
\citep{zhao2018infovae,belghazi2018,gao2018tc}. These works typically require adversarial training.}

Qualitative evaluation of the model includes inspecting samples from prior/variational posterior,
and linearly interpolating in the latent space and evaluating samples from the interpolated latent vector.

\subsection{Tightening the ELBO}\label{tightening}
Ideally we would like the gap between $\log p(x \param \theta) $ and $\ELBO(\theta, \phi \param x)$
to be small. We have seen that the gap is equal to zero if $q(z \given x \param \phi) = p(z \given x \param \theta)$, since
\[ \KL[q(z \given x \param \phi) \Vert p(z \given x \param \theta)] = \log p(x \given \theta) - \ELBO(\theta, \phi \param x). \]
Therefore if we are able to better approximate the true posterior, we can hope to train better
generative models. In this section we briefly review some recent advances that attempt to
tighten the gap.
\subsubsection{Flows}\label{vaeflows}
By working with richer variational families we can better approximate the true posterior.
One way to do so is through \emph{flows}, in which a sample from a simple base density (typically Gaussian) is converted into a sample from a more complex density via a series of invertible transformations, \begin{align*}
\boldz_0 &\sim q(\boldz_0 \given x \param \phi) = \mathcal{N}(\boldsymbol{\mu}(x), \boldsymbol{\sigma^2}(x))\\
\boldz_K &= f_K \circ f_{K-1} \circ \dots \circ f_1(\boldz_0).
\end{align*}
Here $z_0$ is a sample from the variational posterior as before.
Since each $f_k$ is invertible, the log density of $z_K$ is given by the change-of-variables formula
\begin{align*}
\log q_K(\boldz_K \given x \param \phi) &= \log q(\boldz_0 \given x \param \phi) + 
\sum_{k=1}^K \log \Big | \frac{\partial f_k^{-1}}{\partial \boldz_{k}}\Big |  \\
&=\log q(\boldz_0 \given x \param \phi) - 
\sum_{k=1}^K \log \Big | \frac{\partial f_k}{\partial \boldz_{k-1}}\Big | 
\end{align*}
where $ | \frac{\partial f_k}{\partial z_{k-1}}\Big | $ is the absolute value of the determinant 
of the Jacobian of $f_k$. (Here the parameters of $f_k$ have been subsumed into $\phi$).
Then the ELBO using $q_K$ as the variational posterior is
\begin{align*}
\ELBO(\theta, \phi \param x) &= \E_{q_K(\boldz_K \given x \param \phi)} [\log p(x, \boldz_K \param \theta) - \log q_K(\boldz_K \given x \param \phi)] \\
&= \E_{q(\boldz_0 \given x \param \phi)} \Big[\log p(x, \boldz_K \param \theta) - \log q(\boldz_0 \given x \param \phi) + 
\sum_{k=1}^K \log \Big | \frac{\partial f_k}{\partial \boldz_{k-1}}\Big| \Big]\\
\end{align*}
The key to making flows efficient is making the determinant of the Jacobian easy to calculate.
For example, \cite{Rezende2015} use \emph{normalizing flows} where each $f_k$ is parameterized as
\[f_k(\boldz_{k-1}) =  \boldz_{k-1} + \boldu_k h(\boldw_k^\top \boldz_{k-1} + b_k)\]
where $\boldu_k, \boldw_k, b_k$ are learnable parameters (part of $\phi$) and $h(\cdot)$ is a differentiable non-linear function with
derivative $h'(\cdot)$. Then the matrix determinant lemma gives 
\[ \Big| \frac{\partial f_k}{\partial \boldz_{k-1}}\Big| = |1 + \boldu_k^\top (h'(\boldw_k^\top z_{k-1} + b_k)\boldw_k)|, \]
which is very efficient to calculate. Another popular type of flow is the inverse autoregressive flow (IAF) \citep{Kingma2015}. Letting $\boldz_{k, < d}$ refer to the first $d-1$ dimensions of the latent vector at the $k$-th flow step, IAF defines an
invertible transformation via
\[ f_k(\boldz_{k-1}) =  \bmu_k + \bsigma_k \odot \boldz_{k-1}, \]
where $\bmu_{k,d} \in \reals$ and $\bsigma_{k,d} > 0$ (the $d$-th parameters of the above transformation) are the output from a neural network over the previous dimensions $\boldz_{k-1, <d}$,
\[ \bmu_{k,d}, \bsigma_{k,d} = \textrm{MLP}_{k,d}(\boldz_{k-1, <d}) \] 
In this case the Jacobian is given by a triangular matrix, and hence we can efficiently calculate the log determinant,
 \[ \log \Big| \frac{\partial f_k}{\partial \boldz_{k-1}}\Big| = \sum_{d=1}^D \log \bsigma_{k, d}. \]
Note that the MLP's above do not have to be invertible, as long as they only depend on the previous dimensions, and further, we can efficiently calculate $\bmu_k, \bsigma_k$ in a single forward pass using masked autoregressive networks \citep{germain2015}. Hence training remains efficient.

Other types of flows include householder flows \citep{Tomczak2016hf,berg2018}, neural autoregressive flows \citep{huang2018naf}, and \emph{continuous} normalizing flows \citep{chen2018ode,grathwohl2018ffjord}. Continuous flows are particularly interesting in that they instead use neural networks to parameterize a differential equation that 
represents how the layer changes in continuous time.

\subsubsection{Importance Sampling}\label{importancesampling}
Another way to tighten the gap between the ELBO and the log marginal likelihood is through importance sampling~\citep{Burda2015}. 
To illustrate this, first note that, using Jensen's inequality, we have the following:
\begin{align*}
     p(x \param \theta) &= \E_{q(z \given x \param \phi)} \Big[ \frac{p(x, z \param \theta)}{q(z \given x \param \phi)}\Big] \implies 
     \log p(x \param \theta) \ge  \E_{q(z \given x \param \phi)} \Big[\log  \frac{p(x, z \param \theta)}{q(z \given x \param \phi)}\Big].
\end{align*}
It is moreover clear that Jensen's inequality allows us to form an analogous inequality for \textit{any} unbiased estimator of $p(x \param \theta)$, not just $\frac{p(x, z \param \theta)}{q(z \given x \param \phi)}$. In particular, consider the following unbiased estimator, which uses multiple independent samples $z^{(1:K)} = [z^{(1)}, \dots, z^{(K)}]$ from $q(z \given x \param \phi)$:
\[ I_K = \frac{1}{K} \sum_{k=1}^K \frac{p(x, z^{(k)} \param \theta)}{q(z^{(k)} \given x \param \phi)}. \]
Then, using an argument analogous to the one above, and letting $q(z^{(1:K)} \given x \param \phi) = \prod_{k=1}^K q(z^{(k)} \given x \param \phi)$, we have
\[ 
     \log p(x \param \theta) \ge  \E_{q(z^{(1:K)} \given x \param \phi)} 
     \Big[ \log \frac{1}{K} \sum_{k=1}^K \frac{p(x, z^{(k)} \param \theta)}{q(z^{(k)} \given x \param \phi)} \Big] = \E[\log I_K].
     \]

\citet{Burda2015} prove that under mild conditions, we have
\[ \log p(x \param \theta) \ge \E[\log I_K] \ge \E[\log I_{K-1}], \]
and further, that $\lim_{K \rightarrow \infty} \E[ \log I_K] = \log p(x \param \theta)$,
where the convergence is in the strongest (i.e. almost sure) sense.
If $K=1$, we recover the vanilla VAE, and therefore optimizing the above
objective for $K>1$ maximizes a tighter bound.
Under this setup, $q(z \given x \param \phi)$ is viewed as an importance sampling distribution (and $\frac{p(x, z \param \theta)}{q(z \given x \param \phi)}$ the importance weights),
and so this type of approach is known as an \emph{Importance Weighted Autoencoder (IWAE)} \citep{Burda2015}.

In the general case the gradients with respect to $\theta$ and $\phi$ are given by
\begin{align*}
    \nabla_{\theta} \E[\log I_K] &= \E_{q(z^{(1:K)} \given x \param \phi)}\Big[\sum_{k=1}^K w^{(k)} \nabla_{\theta}  \log p(x, z^{(k)} \param \theta) \Big], \\
    \nabla_{\phi} \E[\log I_K] &= \E_{q(z^{(1:K)}  \given x \param \phi)}\Big[\sum_{k=1}^K (\log I_K - w^{(k)}) \nabla_\phi \log q(z^{(k)} \given x \param \phi)  \Big],
\end{align*} 
where 
\[ w^{(k)} = \frac{p(x, z^{(k)})/q(z^{(k)} \given x \param \phi)}{\sum_{j=1}^K p(x, z^{(j)})/q(z^{(j)} \given x \param \phi)}\]
are the normalized importance weights (see \cite{Mnih2016} for the derivation). The expectations here can again be estimated with a single sample from $q(z^{(1:K)} \given x \param \phi)$, which is equivalent to $K$ samples from $q(z \given x \param \phi)$. In the Gaussian case it is still possible to apply the reparameterization trick to the IWAE objective to derive low-variance gradient estimator for $\phi$ (see \cite{Burda2015} for the expression of reparameterized gradient estimator, which is different from score function gradient estimator shown above), and this reparameterized estimator should be used instead. Empirically IWAEs typically outperform VAEs in terms of log marginal likelihood even with a few samples \citep{Burda2015,Mnih2016}. However, the IWAE objective may hinder the training of the inference network
by reducing the signal-to-noise ratio of the gradient estimator \citep{Rainforth2018},
though recent work targets this issue by applying reparameterization twice \citep{tucker2018double}.

These ideas have also been extended to sequential latent variable models with particle filtering \citep{maddison2017fivo,naesseth2018seqmc,le2018}, in which we no longer simply take $q(z^{(1:K)} \given x \param \phi)$ to be the product of $q(z \given x \param \phi)$'s. \cite{cremer2017iwae} and \cite{Domke2018} provide an alternative interpretation of IWAE as optimizing the standard ELBO but with a different variational distribution.

\subsubsection{Amortization Gap}
In VAEs we restrict the variational family to be the class of distributions whose parameters
are obtainable by running  a parameteric model (i.e. inference network) over the input.
This choice allows for fast training/inference but may be too strict of a restriction. (Note that
this is usually not an issue with traditional VI which typically works with variational families that allow
for closed-from expressions for the best $\lambda$ for a given $x$).

Letting $\lambda^\star$ be the best variational parameter for a given data point, i.e.
\[ \lambda^\star = \argmin_\lambda \KL[q(z \param \lambda) \, \Vert \, p(z \given x \param \theta)] \]
we can break down the \emph{inference gap} (the gap between the variational posterior from the inference network and the true posterior) as follows \citep{Cremer2018}
\begin{align*}
    \underbrace{\KL[q(z \given x \param \phi)\,  \Vert \, p(z \given x \param \theta)]}_{\text{inference gap}}  &= \underbrace{\KL[q(z \param \lambda^\star) \, \Vert \, p(z \given x \param \theta)]}_{\text{approximation gap}} + \\ &\,\,\,\,\,\,\,\, \underbrace{\KL[q(z \given x \param \phi) \, \Vert \, p(z \given x \param \theta)] - \KL[q(z  \param \lambda^\star) \, \Vert \, p(z \given x \param \theta)]}_{\text{amortization gap}}
\end{align*} 
Therefore the inference gap consists of two parts: the \emph{approximation gap}, which is the  gap between the true posterior and the best possible variational posterior within $\mathcal{Q}$, and the \emph{amortization gap}, which quantifies the gap between inference network posterior and the best possible variational posterior.

To reduce the approximation gap we can work with richer variational families (e.g. by applying flows),
and to reduce the amortization gap we can better optimize $\lambda$ for each data point. \cite{Cremer2018}
find that both approximation/amortization gaps contribute significantly to the final inference gap,
and there has been some recent work on actively reducing the amortization gap.
For example one could use an inference network to initialize the variational parameters and subsequently run iterative refinement (e.g. gradient ascent) to refine them \citep{Hjelm2016,Krishnan2017,Kim2018}. Alternatively, \cite{Marino2017} utilize meta-learning \citep{Andrychowicz2016} to learn to perform better
inference. Similar ideas have also been explored in \cite{Salakhutdinov2010}, \cite{Cho2013}, \cite{Salimans2015}, and \cite{Pu2017}.

\subsection{Working with Other (e.g. Discrete) Distributions}\label{otherdist}
\subsubsection{Extending the Reparameterization Trick}\label{reparam}
The reparameterization trick is often crucial in training VAEs as it allows us to efficiently
obtain low-variance estimators of the gradient.  
\cite{ruiz2016}, \cite{naesseth2017}, and \cite{figurnov2018implicit} generalize the reparameterization trick to work
with other distributions (e.g. Gamma, Dirichlet). Of particular interest are methods that
extend reparameterization to discrete distributions utilizing the \emph{Gumbel-Max trick}
\citep{Papandreou2011,hazan2012,Maddison2014}. Concretely, suppose $z$
is the one hot representation of a categorical random variable with $K$ categories and unnormalized
scores $\alpha$, i.e.
\[ p(z_k = 1 \param \alpha) = \frac{\alpha_k}{\sum_{i=1}^K \alpha_i}\]
Then we can draw a sample from $p(z \param \alpha)$ by 
adding independent Gumbel noise to the logits and finding the $\argmax$, 
\begin{align*}
    k &= \argmax_{i} [g_i + \log \alpha_i] \\
    z_k &= 1, \,\,\,\, (z_i = 0 \text{ for all } i \ne k)
\end{align*}
where $g_i$'s are independent Gumbel noise.\footnote{We can sample from a Gumbel distribution
by sampling a uniform random variable $u \sim \mathcal{U}[0,1]$ and applying the transformation $g = -\log (-\log u)$.} More succinctly, we have
\[ z = \argmax_{u \in \Delta^{K-1}} (\log \alpha + g)^\top u,\]
where $\Delta^{K-1}$ is the $K$-simplex, $g = [g_1, \dots, g_K]$ is the vector of Gumbel noise, and $\alpha = [\alpha_1, \dots, \alpha_K]$. In this case we have $z \sim p(z \param \alpha)$, and while this shows that we can reparameterize a discrete distribution, we cannot perform gradient-based optimization because the $\argmax$ function has zero gradients almost everywhere.

If we replace the $\argmax$ above with a $\softmax$ and a temperature term $\tau > 0$, i.e.
\begin{align*}
s &= \softmax\Big(\frac{\log \alpha + g}{\tau} \Big) \\
s_k &=  \frac{\exp((\log \alpha_k + g_k)/\tau)}{\sum_{i=1}^K \exp((\log \alpha_i + g_i)/\tau)},
\end{align*}
we say that $s = [s_1, \dots, s_K]$ is drawn from a \emph{Gumbel-Softmax} \citep{Jang2017} or a \emph{Concrete} \citep{Maddison2017} distribution with parameters $\alpha, \tau$.
The Gumbel-Softmax defines a distribution on the simplex $\Delta^{K-1}$, and the density is given by
\[ p(s \param \alpha, \tau) = ( K-1)! \tau^{K-1} \prod_{k=1}^K \Big( \frac{\alpha_k s_k^{-\tau - 1}}{\sum_{j=1}^K \alpha_j s_j^{-\tau} } \Big) .\]
(See \cite{Jang2017} and \cite{Maddison2017} for the derivation).  Notably this distribution is reparameterizable by construction since we can draw a sample by (1)
drawing $i.i.d$ Gumbel noise $g = [g_1, \dots, g_K]$, (2) transforming the noise as $(g_i  + \log \alpha_i)/\tau$, and (3) applying a softmax to the transformed logits. Importantly, unlike the $\argmax$ case, the $\softmax$ relaxation has nonzero gradients.\footnote{Note that replacing the $\argmax$ with the $\softmax$ is equivalent to adding an entropy regularizer to the $\argmax$ problem, i.e.
\[ s = \softmax(\beta) = \argmax_{u \in \Delta^{K-1}} \beta^\top u + H(u),\]
where $H(u) = -\sum_{k=1}^K u_k \log u_k$ is the entropy term. There has been recent work on differentiable relaxations of $\argmax$ with different regularizers, e.g. the \emph{sparsemax} adds a squared penalty \citep{Martins2016,Niculae2018},
\[ \textrm{sparsemax}(\beta) = \argmax_{u \in \Delta^{K-1}} \beta^\top u - \frac{1}{2} \Vert u \Vert_2^2. \]
This is equivalent to a Euclidean projection of $\beta$ onto the simplex, which 
encourages sparsity (since the projection is likely to hit an edge of the simplex).} 
While $s$ is no longer discrete, we can anneal the temperature $\tau \rightarrow 0$ as training
progresses and hope that this will approximate a sample from a discrete distribution.

Going back to VAEs, suppose now $\alpha = \enc(x \param \phi)$ and let 
$q_s(s \given x \param \phi, \tau)$ be the Gumbel-Softmax distribution (note that $\tau$ could in theory also be a function of $x$). If $q_z(z \given x \param \phi)$ is the original categorical distribution, we might hope that the gradient of the ELBO
obtained from the Gumbel-Softmax distribution will approximate the true gradient, i.e.
\begin{align*}
    \nabla_\phi \E_{q_z(z \given x \param \phi, \tau)}\Big[\log \frac{p_{x,z}(x, z \param \theta)}{q_z(z \given x \param \phi)} \Big] & \approx \nabla_\phi \E_{q_s(s \given x \param \phi)}\Big[\log \frac{p_{x,s}(x,s  \param \theta, \tau)}{q_s(s \given x \param \phi)}\Big] \\
    &= \E_{g \sim \text{Gumbel}}\Big[ \nabla_\phi \log \frac{p_\text{x,s}(x, f(g, \alpha, \tau)  \param \theta, \tau)}{q_s( f(g, \alpha, \tau) \given x \param \phi)}\Big],
\end{align*}
where $f$ applies the required softmax transformation to the Gumbel noise $g$ to obtain $s$. Here we have employed subscripts in the density functions to emphasize the random variables over which the distributions are induced: $p_{x,z}(x, z \param \theta) = p_{x|z}(x \given z \param \theta)p_z(z \param \theta)$ is the joint density under the original model while $p_{x,s}(x, s \param \theta, \tau) = p_{x|z}(x \given z= s \param \theta)p_s(s \param \theta, \tau)$ is the density under the new, relaxed model (note that $p_{x,s}$ still makes use of the original likelihood model $p_{x|z}$).\footnote{See the appendix of \cite{Maddison2017} for further discussion on which components can/should be relaxed. In particular, under the KL formulation of the ELBO, there is some debate as to whether
the KL should be calculated between the original categorical distributions (which can be readily calculated in closed-form) or between the relaxed Gumbel-Softmax distributions (which requires sampling).} Unlike the score function gradient estimator, the above estimator is biased and the variance
will diverge to infinity as $\tau \rightarrow 0$. Recent work \citep{Tucker2017,Grathwohl2018} combines
the above estimator with the standard score function estimator to obtain even lower-variance
gradient estimators.\footnote{Some other strategies for reducing the variance of the estimator
include Rao-Blackwellization \citep{Ranganath2014} and sophisticated control variates \citep{Mnih2014,miller2017,deng2018}.} Finally, since we are still using the original likelihood model $p_{x|z}$ to define the relaxed joint distribution $p_{x,s}$,  the above relaxation is
only applicable if the likelihood model $p_{x|z}$, which was originally defined over
a \emph{discrete} latent space $z \in \{0, 1\}^K$ (and the observed data),
is well-defined for a \emph{continuous} latent space $z \in \Delta^{K-1}$. For example if $z$
is a parse tree and $p_{x|z}(x \given z \param \theta)$ uses the parse tree to define a 
complex computational graph (e.g. as in Recurrent Neural Network Grammars \citep{dyer2016rnng}),
then the techniques described above are not applicable because $p_{x|z}(x \given z \param \theta)$ would not make sense for non-discrete $z$.\footnote{However see \cite{choi2018learning} for
an example of relaxing a TreeLSTM to work with ``soft" parse trees in the context of text classification tasks.}

\subsubsection{Non-Gaussian Prior/Posterior Distributions}
Here we briefly review some other distributions that have been considered in the context of VAEs.
\cite{Johnson2016} introduce structured variational autoencoders, which
encode structure in the latent space through probabilistic graphical models. 
Structured VAEs have been used to 
model latent sequences \citep{Chung2015,Fraccaro2016,Serban2017,zaheer2017latent,Krishnan2017b,liu2018hsmm}, graphs \citep{kipf2016vae,jin2018junction},  trees \citep{yin2018structvae,corro2018semi,li2019grammar} and other structured objects \citep{kusner2017grammar,dai2018syntax}. There has also been some work on
extending VAEs to the nonparametric Bayesian setting \citep{Tran2016GP,Goyal2017,naslinick2017,miao2017nvi,singh2017,bodin2017}, and
the von Mises--Fisher distribution is becoming an interesting alternative
to the Gaussian \citep{Guu2017,Davidson2018,xu2018}.

Other classes of distributions that
have been considered include mixture of Gaussians \citep{Dilokthanakul2016}, learned priors \citep{Tomczak2017,huang2017learn}, hierarchical \citep{Son2016,zhao2017hier,park2018}, and discrete \citep{Rolfe2017,vqvae} distributions.

\section{Other Methods}\label{sec:othermethods}
In this section we briefly review some other methods that have been used to train latent variable models.

\subsection{Wake-Sleep Algorithm}
The wake-sleep algorithm \citep{hinton1995ws} is a method for training deep directed graphical models,
and can be thought of as a precursor to variational autoencoders. The wake-sleep algorithm also makes 
use of a \emph{recognition network} which produces an approximate posterior
distribution over the latent variables given the data, i.e. $q(z \given x \param \phi)$. Training proceeds as follows:

\begin{enumerate}
    \item Wake phase: sample a data point $x \sim p_\star(x)$ and latent variable $\hat{z} \sim q(z \given x \param \phi)$. Take a gradient step with respect to $\theta$ to maximize the joint likelihood, 
    \[ \theta^{(i+1)} = \theta^{(i)} + \eta \nabla_{\theta} \log p(x, \hat{z} \param \theta^{(i)})\]
    \item Sleep phase: sample ``phantom" data from the generative model, i.e. $\hat{z} \sim p(z), \hat{x} \sim p(x \given \hat{z} \param \theta)$.  Take a gradient step with respect to $\phi$ to maximize $\hat{z}$, 
    \[ \phi^{(i+1)} = \phi^{(i)} + \eta \nabla_{\phi} \log q(\hat{z} \given \hat{x} \param \phi^{(i)})\] 
\end{enumerate}

It is clear that the wake phase is training the generative model to maximize a Monte Carlo estimate of the expected complete data likelihood under the variational posterior, i.e. $\E_{q(z \given x \param \phi)}[\log p(x, z \param \theta)]$. Hence the wake phase is equivalent to the variational M-step.
In the sleep-phase the recognition network is trained to maximize a Monte Carlo estimate of 
$\E_{p(x, z \param \theta)} [\log q(z \given x \param \phi)]$, which is equivalent to minimizing
$\KL[p(z \given x \param \theta) \Vert q(z \given x \param \phi)]$. Recall that the variational E-step
trains the recognition network to  minimize $\KL[q(z \given x \param \phi) \Vert p(z \given x \param \theta)]$. Therefore the sleep-phase almost corresponds to the variational E-step, but the KL direction is reversed.
We have seen before that in the variational E-step, minimizing  the KL is equivalent to maximizing the ELBO. But this involves an expectation over $q(z \given x \param \phi)$, which is difficult to optimize, especially if the latent variable is not reparameterizable. In contrast, the sleep phase minimizes the other direction
$\KL[p(z \given x \param \theta) \Vert q(z \given x \param \phi)]$ and therefore does not have this issue.

While the wake-sleep algorithm does not maximize a lower bound on the log marginal likelihood, in practice
it performs well \citep{hinton2006,Ba2015,Mnih2016,le2018rws}. \cite{bornschein2015} introduce the reweighted wake-sleep algorithm, which combines importance sampling with wake-sleep to further improve performance.

\subsection{Reversible Neural Generative Models}\label{flows}
Suppose $z$ and $x$ are both continuous, and consider the following generative model
\begin{align*}
    z \sim p_z(z), && x =  f(z \param \theta),
\end{align*}
where $p_z(z)$ is a simple prior (e.g. Gaussian) and $x$ is obtained by applying a deterministic function $f$ to $z$.\footnote{This is in contrast to generative models
considered so far, in which the \emph{parameters} of the distribution $p(x \given z \param \theta)$
are given by applying $f$ to the latent variable $z$.}
If $f$ is invertible, it is possible to evaluate $\log p(x \param \theta)$ using the change-of-variables formula,
\[ \log p(x \param \theta)  = \log  p_z(f^{-1}(x)) + \log \Big|\frac{\partial f^{-1}(x)}{\partial x}\Big|, \]
where $ \Big|\frac{\partial f^{-1}(x)}{\partial x}\Big| $ is the absolute value of the 
determinant of the Jacobian. 
There has been recent work \citep{dinh2015,dinh2017,papamakarios2017, kingma2018} on applying such methods with $f$ parameterized in a way to allow for fast evaluation of the determinant (e.g. using ideas from 
autoregressive density estimation with neural networks \citep{larochelle2011,germain2015} to make the Jacobian a triangular matrix, in which case the log determinant is simply the summation of the log diagonals). Similar transformations have also been applied to variational posteriors to make them more flexible in the context of VAEs \citep{Rezende2015,Kingma2016} (see section~\ref{vaeflows}).
\cite{chen2018ode} and \cite{grathwohl2018ffjord} introduce \emph{continuous normalizing flows}, where the discrete flow steps are mapped to continuous time and a neural network parameterizes the continuous time-dynamics via an ordinary differential equation.

While deep generative models parameterized with reversible neural networks have achieved impressive results in image generation \citep{dinh2017,kingma2018}, they are unfortunately difficult to apply to language since text is typically modeled as a discrete variable. \cite{he2018} circumvent this issue by modeling pretrained word embeddings instead for unsupervised POS tagging and dependency parsing.

\subsection{Generative Adversarial Networks}\label{gan}
Let us consider a similar generative model as above, $z \sim p(z), x =  f(z \param \theta)$,
where $f: \reals^d \to \reals^n$ is differentiable but not necessarily invertible (hence we no longer require that $d = n$). In this case we cannot easily
evaluate $p(x \param \theta)$ using the change-of-variables formula, although the density
can be defined as the partial derivative of the cumulative density function, 
\[ p(x \param \theta) = \frac{\partial^n}{\partial x_1 \dots \partial x_n}\int_{\{z : f(z\param \theta) < x \}} p(z). \]
\emph{Implicit probabilistic models} are generative models like the above which specify a stochastic procedure for generating samples from the model \citep{mohamed2016}. This is in contrast to \emph{prescribed} or \emph{explicit probablistic models} in which the modeler specifies an explicit parameterization of the density $p(x \param \theta)$.\footnote{The distinction between implicit/explicit generative models is therefore based on how the generative model is defined, rather than the distribution itself. For example, if $x \sim \mathcal{N}(\mu, \sigma^2)$, we can implicitly define the model by giving a stochastic procedure for generating the data, i.e.
\[ \epsilon \sim \mathcal{N}(0, 1), \,\,\,\,\,\,\,\,\, x = \mu + \sigma\epsilon,\]
or we can explicitly define it by specifying the density, i.e.
\[ p(x \param \mu, \sigma^2) = \frac{1}{\sqrt{2\pi\sigma^2}}\exp\Big(-\frac{(x-\mu)^2}{2\sigma^2}\Big).\]} In this tutorial we have mostly considered explicit probabilistic models, whose training involved maximizing the log likelihood (or a lower bound on it) directly. Implicit models are instead trained with \emph{likelihood-free} inference, which (roughly) are a class of methods that estimate the density ratio $\frac{p_\star(x)}{p(x \param \theta)}$ or the density difference $p_\star(x) -p(x \param \theta)$
instead of directly working with the likelihood $p(x \param \theta)$ \citep{marin2012,gutmann2012,gutmann2014,bernton2017}.

Generative Adversarial Networks (GAN) \citep{goodfellow2014generative} are deep, implicit generative
models trained with adversarial training where a \emph{discriminator} (or a \emph{critic})
is trained to distinguish between samples from the generative model $p(x \param \theta)$ versus the true data distribution $p_\star(x)$. The generative model is trained ``adversarially" to fool the discriminator.

Formally, let $D: \mathcal{X} \to (0, 1)$ be the discriminator (typically parameterized as 
a neural network) that maps the data  $x \in \mathcal{X}$ to the interval $(0,1)$. The discriminator parameters $\psi$ and the generative model parameters $\theta$ are trained based on the following
minimax objective:

\[ \min_\theta \max_\psi \E_{p_\star(x)}[\log D(x \param \psi)] + \E_{p(x \param \theta)}[\log ( 1 - D(x \param \psi))], \]

where the expectations are approximated with samples from $p_\star(x)$ and $p(x \param \theta)$. Interpreting $D(x \param \psi)$ as the probability the discriminator assigns to the event that $x$ is sampled from the true data distribution, we can see that 
the discriminator training objective is the standard log likelihood objective from 
logistic regression. The generative model is trained to minimize the probability that the discriminator will correctly classify the generated sample as fake.\footnote{In practice the generator is trained to maximize $\E_{p(x\param\theta)}[- \log D(x \param \psi)]$ as this provides stronger gradient signals early in training.} If the observed domain $\mathcal{X}$ is continuous (e.g. dequantized images) then we can backpropagate the gradients from the discriminator to the
generative model. For text this is generally not the case and thus we must resort to
other techniques.

\cite{goodfellow2014generative} show that the above objective
approximately minimizes the Jensen-Shannon divergence between the data/model distributions, which is
given by
\[ \JS[p_\star(x) \Vert p(x \param \theta)] = \frac{1}{2}\Big(\KL\Big[p_\star(x) \Big\Vert \frac{p_\star(x) + p(x \param \theta)}{2} \Big] + \KL\Big[p(x \param \theta) \Big\Vert \frac{p_\star(x) + p(x \param \theta)}{2} \Big]  \Big). \]
This is notably different from the usual maximum likelihood training which minimizes $\KL[p_\star(x) \Vert p(x \param \theta)]$.
The GAN objective has been generalized to (approximately) minimize other measures, 
such as $f$-divergences \citep{nowozin2016f} and the Wasserstein distance \citep{arjovsky2017wasserstein,tolstikhin2017wasserstein}, which requires different
parameterizations of the discriminator $\psi$.

Empirically GANs perform remarkably well and are able to generate impressive-looking
images \citep{Radford2016,salimans2016,zhang2017stack,miyato2018,Karras2018prog,brock2018largegan}. While a rich
research program has formed around GANs (and implicit
models in general) over the past few years,\footnote{For example see \url{https://github.com/hindupuravinash/the-gan-zoo} for a list of GAN variants.}
their applications to text modeling has been somewhat limited by the difficulty
associated with optimizing the GAN objective when the output space is discrete.

\subsubsection{Generative Adversarial Networks for Text}\label{textgan}
Training the discriminator is straightforward with gradient-based methods.
For the generative model if 
the domain $\mathcal{X}$ is continuous and $D(x \param \psi)$ is differentiable with 
respect to $x$, then we can use the reparameterization trick to obtain low variance gradient estimators
\begin{align*}
    \nabla_\theta \E_{p(x \param \theta)}[\log(1 - D(x \param \psi)] &= \nabla_\theta \E_{p(z)} [\log(1 - D(f(z \param \theta) \param \psi)]  \\
    &= \E_{p(z)}[\nabla_\theta \log (1-D(f(z \param \theta) \param \psi))] \\
    &= \E_{p(z)}\Big[ - \frac{1}{1-D(x \param \psi)}  \nabla_x D(x \param \psi)\frac{\partial x}{\partial \theta}  \Big],
\end{align*} 
where the expectation is approximated with Monte Carlo samples. Notice that this requires
the gradient of $D(x \param \psi)$ with respect to $x$, which is multiplied with the Jacobian
$\frac{\partial x}{\partial \theta}$ to obtain the gradient with respect to $\theta$.\footnote{Note that automatic differentiation packages never actually instantiate the full Jacobian, but we have written it explicitly here for illustrative purposes.}

If $\mathcal{X}$ is discrete however, we run into several issues.
First, we cannot even have $x = f(z \param \theta)$ since a deterministic function that maps from a continuous $z$ to a discrete $x$ will either be uninteresting (e.g. a constant function)
or badly behaved (e.g. an argmax, which is differentiable almost everywhere but has zero gradients). A tempting strategy in this case is to define the likelihood $p(x \given z \param \theta)$ whose \emph{parameters} are the output from $f$, e.g. for a single token $x$
\[ p(x \given z \param \theta) = \pi_x \,\,\,\,\,\,\,\,\,\,\,\,\,\,\,\, \pi = f(z \param \theta) = \softmax(\MLP(z \param \theta)).\]
Note that this is now an \emph{explicit generative model}, much like the models we have considered in previous sections. Then the score function gradient estimator is
\begin{align*}
   \E_{p(x \param \theta)}[\log(1 - D(x \param \psi)) \nabla_\theta \log p(x \param \theta)].
\end{align*}
Unfortunately we run into another issue here because $\log p(x \param \theta)$ was assumed to
be intractable in the first place! Researchers have proposed to various ways to mitigate this problem.
\paragraph{Relaxing the Discrete Space} Several works \citep{Gulrajani2017wgan,Rajeswar2017,Press2017} relax the discrete space into a continuous space, for example by using the softmax function instead of
the one-hot representation of $x$,
\[ x = f(z \param \theta) = \softmax(\MLP(z \param \theta)). \]
If $x$ is a sequence of discrete symbols, then the output would be a $\softmax$-ed vector for each time step, i.e. a $T \times V$ dense matrix where $T$ is the number of tokens and $V$ is the vocabulary size. For example, we may have separate MLPs for each position, 
\[
    x = [x_1, \dots x_T]    \,\,\,\,\,\,\,\,\,\,\,\,\,\,\,\,  x_i = f_i(z \param \theta) = \softmax(\textrm{MLP}_i(z \param \theta)).
\]
(Other parameterizations are possible, for example one might use an RNN or a transposed convolutional network to share parameters.)

In this relaxed setup, as was the case with images it is straightforward to apply the reparameterization trick to obtain gradients for the generative model. At first this strategy might seem hopeless 
since the discriminator can easily discriminate between a discrete sample $x \sim p_\star(x)$
and a generated (continuous) sample $x \sim p(x \param \theta)$. However in practice this strategy does work to an extent, especially if one instead implicitly minimizes the Wasserstein distance \citep{arjovsky2017wasserstein}. 
\cite{Gulrajani2017wgan} note that training might be feasible in this regime since 
under mild conditions, the Wasserstein distance 
is finite and differentiable almost everywhere (i.e. even if the supports between the two distributions do not overlap).  For actual sample generation it is common practice to replace the $\softmax$ with an $\argmax$ in these models. 

Another way to ``relax"  the discrete space to learn an auxiliary continuous space in which to generate / discriminate \citep{zhao2017adversarially,subramanian2018gan}. For example, \cite{zhao2017adversarially} use the adversarial autoencoder framework \citep{Makhzani2015} to jointly learn (1) a sentence-level autoencoder which encodes into and decodes from a continuous space, and (2) a GAN which learns to generate/discriminate in the same continuous space.
\cite{tolstikhin2017wasserstein} show that this style of training approximately minimizes the Wasserstein distance between the model/data distributions, and \cite{gu2018dialog} apply similar techniques to model dialogue responses.

\paragraph{Non-Latent Variable Models} A simple alternative is to model $p(x \param \theta)$
without an explicit latent variable, e.g. as an RNN language model \citep{Yu2017,lin2017adversarial,guo2018leakgan}. Then the
score function gradient estimator is straightforward to calculate, and this setup
resembles sequence-level training of deep models \citep{ranzato2016seq} where the reward comes from a learned
discriminator. Such methods usually require initializing with a pretrained  language model (trained with maximum likelihood), in addition
to other techniques from reinforcement learning.
\cite{Li2017}, \cite{wu2017adnmt}, and \cite{yang2018nmtgan} extend this setup to the conditional case for neural machine translation and dialogue modeling. Some works instead perform 
adversarial training on the hidden states of the generative model to avoid optimization challenges associated with score function gradient estimators \citep{lamb2016,xu2017embgan}.
We emphasize that these models
are not latent variable models in the sense we have been employing in this tutorial. Instead they should be viewed as language models optimized with adversarial training.
The motivation for this setting is avoiding \emph{exposure bias} \citep{bengio2015sampling,ranzato2016seq,Wiseman2016}, which describes
the mismatch between training (which uses ground-truth history) and generation
(which uses model-generated history). However, whether adversarially trained models are able to meaningfully mitigate exposure bias is an open question \citep{tevet2018gan,caccia2018gan}.

We conclude this section by noting that training implicit models for text is very much an open problem. Evaluation of implicit models itself is difficult, as unlike explicit generative models it is not possible to tractably estimate the log likelihood. Even if log likelihood is estimable, \cite{Theis2016} show that 
models that achieve good log likelihoods do not necessarily generate good samples.
\cite{cifka2018} and \cite{sem2018ganeval} explore a variety of deep generative models of language across various metrics and find that different models do well on different metrics.

\section{Discussion: Role of Latent Variables in Deep Learning}\label{sec:discussion}

Deep latent variable models, which combine the composability and interpretability of graphical models with the flexible modeling capabilities of deep networks, are an exciting area of research. It is nonetheless worth discussing again \emph{why} we would want to use latent variable models in the first place, especially given that from a pure performance standpoint, models that do not formally make use of latent variables such as LSTMs and transformer networks are incredibly effective  \citep{vaswani2017attention,melis2018sota,merity2018regularizing}. Even from the perspective of learning interesting/meaningful structures (e.g. topic modeling, word alignment, unsupervised tagging), one may argue that these meaningful representations can be implicitly captured within the hidden layers of a deep network.
Indeed, the recent, astonishing success of pretrained (non-latent variable) language models as
generic feature extractors \citep{peters2018deep,radford2018gen,devlin2018bert} suggests that deep networks can capture significant amount of linguistic knowledge  without the explicit use of latent variables.

These are valid points, and it is entirely possible that deterministic deep networks are ``all you need". However we think that this is unlikely to be the case.
For one, latent variable models \emph{can} surpass the performance of deterministic deep networks if properly optimized. For example, deterministic (soft) attention \citep{Bahdanau2015} was generally thought to outperform latent variable (hard) attention \citep{Xu2015}, until recent work showed that when properly
optimized (albeit with a more expensive training procedure), latent variable attention
can outperform deterministic attention \citep{deng2018,shankar2018,wu2018}. Similarly, language models equipped with a latent variable per token~\citep{yang2018breaking} are currently the state-of-the-art in language modeling.
Framing certain aspects of a network as essentially approximating a latent variable objective often yields valuable insights and new avenues for further work. 
Notably,  the interpretation of dropout \citep{Hinton2012,Srivastava2014} as optimizing a latent variable objective has led to rich extensions and improvements \citep{kingma2015dropout,gal2016dropout,gal2016dropout2,ma2017dropout,melis2018pushing}.

Finally, latent variable modeling gives us a declarative language with which to explicitly inject both \emph{inductive bias} and domain-specific \textit{constraints} into models. Inductive bias, or the inherent ``preferences" of a model (or learning algorithm), is crucial for learning and generalization. It can help mitigate against model misspecification, allow for data-efficient learning, and through a carefully crafted generative model, enable interesting structures to emerge. Furthermore, if we have constraints on the representations learned by a model, such as that they represent a valid parse tree, or be interpretable, or allow for controlling the model's predictions, we can enforce these constraints in a principled way through latent variables.

\section{Conclusion}\label{sec:conclusion}
This tutorial has focused on learning deep latent variable models of text, in particular, models that can be expressed as directed graphical models. We have reviewed some archetypical models of text including their applications and explored learning such models through maximizing the log marginal likelihood or a lower bound on it. We have  devoted a significant portion of the tutorial to amortized variational inference, a key technique for learning deep latent variable models in which a separate inference network is trained to perform approximate posterior inference. 

We again emphasize that many important areas were not covered in the tutorial. Notably we have not covered undirected graphical models, posterior inference through Markov chain Monte Carlo, spectral learning of latent variable models, and non-likelihood-based approaches. These areas are all important on their own, and the intersection of latent variable models and deep learning remains an exciting avenue for much future work.

\subsection*{Acknowledgments}
We thank Justin Chiu, Yuntian Deng, and Andr{\'e} Martins for providing valuable
feedback. We also thank the developers of PyTorch \citep{paszke2017automatic} and Pyro \citep{pyro2018}, upon which the tutorial code is based. YK is
supported by a Google Fellowship.
\small
\bibliography{master}
\bibliographystyle{acl_natbib_nourl}
\end{document}